%% file: main.tex
\documentclass[authoryear,final,5p,times,twocolumn]{elsarticle}
\usepackage{cite}
\usepackage{amsmath,amssymb,amsfonts}
\usepackage{algorithmic}
\usepackage{graphicx}
\usepackage{booktabs}
\usepackage{subcaption}
\usepackage{multirow} 
\usepackage{algorithm,algorithmic}
\usepackage{hyperref}
\usepackage{array}
\usepackage{textcomp}
\usepackage{tikz}
\usepackage{xcolor}
\usepackage{graphicx}
\usepackage{csquotes}
\usepackage[table]{xcolor}
\newcommand{\legendbox}[1]{\tikz[baseline=0.5ex]\fill[#1] (0,0) rectangle (0.3cm,0.3cm);}
\definecolor{color1}{RGB}{255,0,0}
\definecolor{color2}{RGB}{0,255,0}
\definecolor{color3}{RGB}{0,0,255}
\definecolor{color4}{RGB}{255,255,0}
\definecolor{color5}{RGB}{255,0,255}
\newcommand{\review}[1]{\textcolor{black}{#1}}  
\usepackage{amssymb}
\usepackage{amsmath}
\newcommand{\INPUT}{\item[\textbf{Input:}]}
\newcommand{\OUTPUT}{\item[\textbf{Output:}]}
\journal{Medical Image Analysis}

\begin{document}

\begin{frontmatter}

\title{Adapting SAM to Nuclei Instance Segmentation and Classification via Cooperative Fine-Grained Refinement}

\author[label1]{Jingze Su}
\author[label1]{Tianle Zhu}
\author[label1]{Jiaxin Cai}
\author[label1]{Zhiyi Wang}
\author[label1]{Qi Li}
\author[label1]{Xiao Zhang}
\author[label3]{Tong Tong}
\author[label2]{Shu Wang \corref{correspondingauthor}}
\author[label1]{Wenxi Liu \corref{correspondingauthor}}

\affiliation[label1]{organization={College of Computer and Data Science},
            addressline={Fuzhou University}, 
            city={Fuzhou},
            country={China}}
\affiliation[label2]{organization={School of Mechanical Engineering and Automation},
            addressline={Fuzhou University}, 
            city={Fuzhou},
            country={China}}
\affiliation[label3]{organization={College of Physics and Information Engineering},
            addressline={Fuzhou University}, 
            city={Fuzhou},
            country={China}}

\cortext[correspondingauthor]{These authors are co-corresponding authors.}

\begin{abstract}
Nuclei instance segmentation is critical in computational pathology for cancer diagnosis and prognosis. Recently, the Segment Anything Model has demonstrated exceptional performance in various segmentation tasks, leveraging its rich priors and powerful global context modeling capabilities derived from large-scale pre-training on natural images. However, directly applying SAM to the medical imaging domain faces significant limitations: it lacks sufficient perception of the local structural features that are crucial for nuclei segmentation, and full fine-tuning for downstream tasks requires substantial computational costs. To efficiently transfer SAM's robust prior knowledge to nuclei instance segmentation while supplementing its task-aware local perception, we propose a parameter-efficient fine-tuning framework, named Cooperative Fine-Grained Refinement of SAM, consisting of three core components: 1) a Multi-scale Adaptive Local-aware Adapter, which enables effective capability transfer by augmenting the frozen SAM backbone with minimal parameters and instilling a powerful perception of local structures through dynamically generated, multi-scale convolutional kernels; 2) a Hierarchical Modulated Fusion Module, which dynamically aggregates multi-level encoder features to preserve fine-grained spatial details; and 3) a Boundary-Guided Mask Refinement, which integrates multi-context boundary cues with semantic features through explicit supervision, producing a boundary-focused signal to refine initial mask predictions for sharper delineation. These three components work cooperatively to enhance local perception, preserve spatial details, and refine boundaries, enabling SAM to perform accurate nuclei instance segmentation directly. Extensive experiments on three challenging benchmarks demonstrate that our method achieves state-of-the-art performance, outperforming existing methods while requiring significantly fewer trainable parameters.
\end{abstract}

\begin{keyword}
Nuclei Instance Segmentation, Segment Anything Model, Parameter-Efficient Fine-Tuning, Computational Pathology.
\end{keyword}

\end{frontmatter}

\section{Introduction}

With the continuous advancement of whole slide imaging (WSI) generation and scanning technologies, pathological image analysis tasks such as nucleus segmentation and classification have gained increasing importance in cancer diagnosis \citep{graham2019hover}. These tasks form the foundation for morphological quantification and tumor grading assessments. By providing essential quantitative characterization of nuclear morphology and architecture, they play a critical role in cancer diagnosis, treatment planning, and survival analysis, as supported by extensive previous studies \citep{sirinukunwattana2018novel, lu2018nuclear}.

\input{fft_motivation}

Given the vast number of cells in a WSI, large-scale cellular analysis has traditionally been time-consuming, labor-intensive, and highly dependent on pathologists' expertise, making consistent manual assessment particularly challenging \citep{elmore2015diagnostic}. With the recent advances in deep learning, convolutional neural network (CNN)-based frameworks have emerged as a powerful alternative, offering automated, efficient, and reproducible analysis. As a result, CNN-based methods have been widely adopted in recent years for nucleus segmentation and classification tasks \citep{graham2019hover, schmidt2018cell,doan2022sonnet}. Their ability to extract local features while preserving fine-grained details helps mitigate common challenges in histopathology, such as faint nuclear boundaries, staining variations, and certain degrees of morphological diversity. However, the inherently limited receptive field of convolutional layers remains a fundamental bottleneck, which prevents them from effectively capturing global context and long-range spatial relationships \citep{dosovitskiy2020image, horst2024cellvit}. Nuclear overlapping, clustering, and extreme morphological diversity continue to pose a significant challenge to the development of robust segmentation algorithms \citep{ilyas2022tsfd}.

To overcome these limitations, recent efforts have turned to foundational vision models that excel in capturing global context and long-range dependencies. Among them, the Segment Anything Model (SAM) \citep{kirillov2023segment} have demonstrated remarkable performance and strong generalization capabilities in natural image segmentation \citep{kirillov2023segment}, prompting researchers to explore its adaptation to nucleus instance segmentation \citep{deng2025segment}. Although SAM excels in long-range dependency modeling and cross-domain generalization, its application to nucleus segmentation still faces several limitations: (1) The full fine-tuning strategy is computationally expensive and necessitates storing a huge number of parameters \citep{hu2022lora}. Moreover, it is prone to overfitting and unable to learn generalized representations due to the limited amount of annotated nucleus data \citep{wu2025medical}; (2) While the Transformer backbone in SAM effectively captures global context, its reliance on self-attention mechanisms is inherently less suited for perceiving fine local details and boundary features, leading to degraded performance in regions with faint or densely clustered nuclei, as evidenced by the results in Fig. \ref{fig:motivation}; (3) Most existing approaches \citep{horst2024cellvit,nam2024instasam} rely on a multi-stage pipeline of semantic segmentation followed by complex post-processing (e.g., watershed-based methods inspired by HoVer-Net \citep{graham2019hover}) to separate instances. Such pipelines are often fragile and highly sensitive to parameter tuning, introducing additional error.

To address these challenges, we propose Cooperative Fine-grained Refinement of SAM (CFR-SAM), a parameter-efficient fine-tuning framework. \review{Our framework consists of two stages. The first stage is a prompt learning module that automatically generates high-quality point prompts by detecting nuclei centroids. The second stage, which embodies our core contributions, takes these prompts as input and performs fine-grained instance segmentation using an adapted SAM.} Our strategy is built upon an end-to-end pipeline that directly leverages SAM's mask decoder with automated point prompts, thereby eliminating complex post-processing. However, to fully realize the potential of this pipeline for fine-grained nuclei segmentation, we need to enhance it against two inherent characteristics: (1) its underutilization of shallow encoder features, which contain rich detailed information (e.g., shape and edges) \citep{hatamizadeh2022unetr}; and (2) its reliance on low-resolution predictions followed by upsampling, which leads to inferior mask detail recovery compared to modern multi-stage decoding architectures \citep{ma2024segment,horst2024cellvit}.
To overcome these intertwined limitations in a cooperative manner, CFR-SAM introduces three core components that work in concert.

First of all, we introduce a Multi-scale Adaptive Local-aware Adapter (MALA), a parameter-efficient fine-tuning (PEFT) module that augments the frozen SAM backbone. While PEFT strategies like Adapter \citep{houlsby2019parameter} or LoRA \citep{hu2022lora} enable efficient adaptation of SAM to medical images \citep{wu2025medical, zhang2023customized}, they typically adjust features only along the channel dimension and lack explicit modeling of local contextual information, which is crucial for nuclear segmentation. To overcome this, MALA incorporates multi-scale convolutional operations within a low-rank space to enhance the local receptive field of each token. This design allows the model to maintain the global modeling strengths of the Transformer while significantly strengthening its perception of local features such as nuclear boundaries, morphology, and size. To handle the substantial variations in nuclear density and scale, MALA adaptively generates pixel-wise convolutional kernels conditioned on the input, thereby matching the optimal receptive field for different cellular regions and effectively extracting region-specific details.
Thus, the local features enhanced by MALA provide a superior foundation for the subsequent modules. 

Next, to directly compensate for the decoder's specific shortcomings, we introduce two dedicated components, the Hierarchical Modulated Fusion Module (HMFM) and the Boundary-Guided Mask Refinement (BGMR), that operate cooperatively. Specifically, HMFM tackles the underutilization of shallow features by dynamically aggregating and modulating features from all encoder levels, including those already enriched by MALA. This ensures that fine-grained spatial details are preserved and directly contribute to the decoding process. Building upon this enhanced feature representation, BGMR addresses the boundary ambiguity problem, which leverages the Multi-Context Boundary Enhancement (MCBE) component to extract explicit, multi-scale boundary cues from the input image. These cues are then aligned and fused with the upsampled, semantically rich features (which combine output from HMFM and the mask decoder) via a Boundary-Guided Feature Fusion (BGFF) module. This fusion process is supervised by an explicit boundary loss, compelling the model to generate a boundary-focused feature representation, which is subsequently used to refine the mask predictions for sharper delineation and improved structural consistency.

\review{To sum up, guided by a task-driven analysis of nuclei segmentation challenges, we identify three inherent limitations in SAM's architecture that hinder its direct application. These problems are independent and occur at different processing stages, thus requiring separate compensatory mechanisms that work in sequence. First, SAM's backbone lacks local detail sensitivity, and MALA addresses this at the feature level. Second, SAM's decoder ignores shallow features, and HMFM handles this at the fusion level. Third, SAM's low-resolution output blurs boundaries, and BGMR compensates for this at the refinement level. These three problems are independent and require separate solutions that work in sequence: MALA provides locally-aware features for HMFM, and HMFM's enriched features then support BGMR's refinement. This layered design is not arbitrary but a necessary way to systematically overcome SAM's shortcomings. Together with the essential prompt learning stage that generates accurate nuclei centroids, CFR-SAM achieves end-to-end, fine-grained instance segmentation without relying on fragile post-processing.} For evaluation, the proposed method achieves state-of-the-art performance on three challenging benchmarks. Notably, on PanNuke, it outperforms the previous leading method (which required full fine-tuning) by 1.62\% in bPQ and 1.34\% in mPQ, while utilizing only 10.6\% of the trainable parameters.

Overall, our contributions are summarized as follows:
\begin{itemize}
\item To address the limitations of full fine-tuning and SAM's lack of local feature sensitivity, we propose the Multi-scale Adaptive Local-aware Adapter (MALA). This parameter-efficient module augments the frozen SAM backbone by injecting multi-scale local features via low-rank adaptation, significantly enhancing nuclear structure perception with minimal overhead.

\item To overcome the mask decoder's underutilization of shallow features, we introduce the Hierarchical Modulated Fusion Module (HMFM). It dynamically aggregates and modulates multi-level encoder features through a channel-wise attention mechanism, ensuring the preservation of fine-grained spatial details for more precise segmentation.

\item To resolve the boundary ambiguity inherent in the mask decoder's upsampling process, we design the Boundary-Guided Mask Refinement (BGMR). It explicitly incorporates boundary supervision, fusing multi-context cues with semantic features to sharply refine mask predictions, producing high-quality instance masks without complex post-processing.

\item CFR-SAM achieves state-of-the-art performance on three challenging benchmarks and it outperforms the previous full fine-tuning methods while utilizing much less trainable parameters.
\end{itemize}

\section{Related Work}

\subsection{Instance Segmentation of Nuclei}
In recent years, nucleus instance segmentation has emerged as a key task in computational pathology, with the goal of accurately identifying and segmenting each nucleus in microscopic images. Traditional image processing methods, such as the watershed algorithm and level set methods, have been used for nucleus segmentation. However, these methods often rely on handcrafted features, making them sensitive to image quality and noise, and limiting their ability to generalize across different datasets \citep{wienert2012detection, veta2013automatic,corbetta2025multi}. 

With the development of deep learning, many methods have significantly improved segmentation performance \citep{he2017mask,graham2019hover,schmidt2018cell,chen2023cpp,horst2024cellvit,li2025nuhtc,ding2025artificial}. To address the common problems of nuclear crowding and occlusion in whole slide images, some studies have begun to incorporate morphological priors, first predicting nuclear semantic segmentation and morphological features, and then extracting individual instances through refined post-processing. Specifically, HoVer-Net \citep{graham2019hover} captures nuclear boundary transitions by computing the gradients of horizontal and vertical distance maps, while STARDIST \citep{schmidt2018cell} and its extension CPP-Net \citep{chen2023cpp} predict the distance from each foreground pixel to the instance boundary along predefined directions. Although these methods have achieved excellent performance, they still face considerable difficulties when dealing with highly heterogeneous nuclei or low-contrast images due to the limited receptive field of convolutional layers. To address these limitations, we propose a novel parameter-efficient paradigm that effectively enhances the model's ability to capture nuclear morphology and contours, highlighting the advantage of leveraging fine-grained details for nuclei segmentation.

\subsection{Utilization of SAM for Medical Image Segmentation}
In recent years, the powerful capability of SAM has garnered significant research interest in the field of medical image segmentation\citep{jiao2026foundation}. Initially proposed by Kirillov et al.\citep{kirillov2023segment}, SAM introduced a prompt-based segmentation mechanism that demonstrated impressive zero-shot generalization ability, marking a breakthrough in natural image segmentation. Following its success in natural image tasks, several studies have explored SAM's direct application to medical image segmentation. However, early investigations (e.g., \citep{deng2025segment} and \citep{huang2024segment}) revealed that while SAM performs well on certain tasks, it faces significant limitations in medical imaging tasks such as nucleus, organ, and lesion segmentation. These limitations include weak domain adaptability, insufficient sensitivity to fine structural details, and a heavy reliance on manual prompt inputs.

To address these challenges and further enhance SAM's performance in medical imaging, researchers have proposed various fine-tuning and extension methods. For instance, SPPNet employs a \enquote{one-prompt-all-nuclei} strategy to fine-tune a distilled lightweight variant of SAM for nuclear semantic segmentation\citep{xu2023sppnet}. Although this method improves segmentation outcomes, it still relies on manual prompting and fails to provide nuclear instance information. CellViT leverages SAM's pre-trained image encoder as the backbone within a Vision Transformer-based U-Net architecture for nuclear instance segmentation\citep{horst2024cellvit}. A limitation of this approach is that it does not fully exploit the knowledge in SAM's integrated architecture (e.g., its prompt encoder and mask decoder). Furthermore, it relies on elaborate post-processing techniques to derive instance segmentation results from semantic masks. In this work, our method fully utilizes the pre-trained knowledge of the entire SAM architecture. By employing automatically predicted nuclear centroids as point prompts, our approach achieves nuclear instance segmentation without relying on manual prompts or elaborate post-processing.

\subsection{Parameter-Efficient Fine-Tuning (PEFT) for SAM}
The adaptation of SAM to medical image segmentation has increasingly used PEFT strategies. Since its introduction, SAM has garnered significant attention owing to its substantial parameter size and strong generalization capabilities. However, full-parameter fine-tuning entails notable challenges such as high computational expense and potential overfitting. These limitations have motivated researchers to adopt PEFT techniques, which are originally popularized in natural language processing, for vision models like SAM. For example, Zhang et al. \citep{zhang2023customized} applied Low-Rank Adaptation (LoRA), a method introduced by Hu et al. \citep{hu2022lora}, to fine-tune SAM’s image encoder using low-rank matrix decomposition, substantially reducing the number of trainable parameters. Similarly, adapter-based approaches have been proposed, incorporating lightweight modules into Transformer layers to facilitate domain adaptation \citep{wu2025medical}.

\review{To further alleviate the heavy reliance on extensive annotated data and computational resources, recent studies have begun to explore more refined parameter- and data-efficient learning strategies tailored for SAM. On the parameter efficiency front, approaches have introduced task-specific efficient tuning modules \citep{paranjape2024adaptivesam} and SVD-based lightweight approximations \citep{paranjape2024s} to significantly reduce computational overhead during adaptation. Concurrently, to address data scarcity and enhance model flexibility, other works have proposed conditional tuning mechanisms for few-shot adaptation \citep{xiao2024cat} and integrated text-driven paradigms for more universal segmentation \citep{koleilat2024medclip}. }

\review{Building on these ideas, recent research has increasingly focused on integrating prompt learning within the PEFT framework to tackle the dual challenge of limited annotations and fine-grained structure segmentation. For instance, \citet{konwer2025enhancing,wang2024masked} developed an annotation-efficient prompting strategy that generates prompts in a fully unsupervised fashion, capturing essential semantic and spatial information to guide SAM in low-data scenarios. Similarly, \citet{spiegler2025textsam} proposed a text prompt learning approach that, combined with LoRA, tunes less than 1\% of SAM's parameters for automatic pancreatic tumor segmentation in ultrasound images. }

\review{The integration of prompt learning with PEFT is particularly advantageous in medical imaging, where annotated data are scarce. Unlike conventional fine-tuning that relies on abundant pixel-level labels, prompt learning leverages task-specific cues, such as semantic prompts (e.g., text embeddings) or spatial prompts (e.g., points or boxes), to steer SAM's pre-trained knowledge. This approach offers two key benefits in data-limited scenarios: semantic prompts encode high-level priors that guide attention to relevant features, compensating for limited visual examples \citep{koleilat2024medclip, spiegler2025textsam}; spatial prompts derived from sparse annotations (e.g., centroid points) enable instance segmentation without dense masks \citep{konwer2025enhancing}. By maximizing the utility of each labeled sample, prompt learning enables robust adaptation to specialized medical tasks where large-scale, fine-grained labels are impractical. Collectively, these studies demonstrate that combining prompt learning with PEFT effectively harnesses semantic cues to mitigate data scarcity while maintaining parameter efficiency.}

Despite these advances, existing PEFT methods continue to face challenges such as limited generalization across heterogeneous datasets and insufficient sensitivity to fine-grained anatomical structures. In particular, when applied to nucleus instance segmentation, these methods often struggle to capture multi-scale local features effectively and to delineate precise boundaries, indicating considerable room for improvement.

\subsection{Lightweight Architectures for Nuclei Segmentation}
\review{In the broader context of efficient medical image analysis, alongside PEFT techniques that focus on adapting foundation models, an orthogonal line of research strives for efficiency through innovative architectural design. Representative works, such as NuLite \citep{tommasino2026nulite} and HoVer-UNet \citep{tommasino2024hover,wu2025segnet}, achieve significant reductions in absolute parameter counts and inference costs by constructing lightweight, task-specific networks while maintaining state-of-the-art performance. Distinct from these lightweight architectures that prioritize absolute inference speed, our efficiency claims are inherently centered on the parameter efficiency during the fine-tuning phase. Rather than competing on inference latency, our core objective is to provide a synergistic PEFT framework that, with minimal parameter updates, effectively stimulates and transfers the robust generalization capabilities of universal foundation models to data-scarce medical imaging scenarios.}

\input{Pipeline}

\section{Proposed Method}

\subsection{Preliminary: SAM architecture}
\review{To establish the foundation for our method, we briefly review the architecture of SAM. As a promptable segmentation model, SAM comprises three main components: an image encoder, a prompt encoder, and a mask decoder. The image encoder is based on a standard Vision Transformer (ViT) pre-trained by MAE. Given an input image $I \in \mathbb{R}^{3 \times H \times W}$, it produces a feature embedding $F \in \mathbb{R}^{C \times H/16 \times W/16}$ that is downsampled by a factor of 16. Meanwhile, the prompt encoder encodes various prompts (e.g., points, boxes, masks) into prompt embeddings. The mask decoder is a tailored Transformer decoder block. It employs two-way cross-attention to model interactions between the prompt and image embeddings. A learnable output token, processed through this decoder, is then mapped by an MLP to form a dynamic linear classifier, which finally predicts the target mask. It is worth noting that SAM generates low-resolution masks followed by upsampling to the original image size. }

\subsection{Overview}
As illustrated in Fig. \ref{fig:pipeline}, our model achieves automated nuclei instance segmentation through two principal stages: Prompt Learning and SAM Adaptation.

In the first stage, a lightweight convolutional network, termed the Prompt Learner, predicts the centroid coordinates and category labels of all nuclei. These predicted centroid coordinates are then fed into second stage as a set of point prompts for the subsequent segmentation. Formally, given an input image $I \in \mathbb{R}^{3 \times H \times W}$, the Prompt Learner (Sec. \ref{method b}) produces a set of prompt tuples $\mathcal{D} = \{d_i\}_{i=1}^{N}$, where each $d_i =(p_i,c_i)$ comprises a point prompt $p_i=(x_i,y_i)$ denoting a nucleus centroid, and its category label $c_i \in \mathbb{C}$. Here, $N$ and $\mathbb{C}$ represent the predicted number of nuclei in the image and the set of nuclei category, respectively.

In the second stage, the point prompts $\mathcal{P} = \{p_i | (p_i,c_i) \in \mathcal{D}\}$ guide our parameter-efficient adaptation of SAM, namely CFR-SAM, to perform instance segmentation. Specifically, the image $I$ and prompts $\mathcal{P}$ are fed into CFR-SAM, whose encoders are adapted by the MALA module (Sec. \ref{method c}). Multi-scale features are then aggregated by the HMFM (Sec. \ref{method d}), and initial mask predictions are progressively refined by the BGMR module (Sec. \ref{method e}). This refinement process incorporates fine-grained edge information to enhance boundary clarity. The final output is a set of refined instance masks $\{\mathcal{M}^{\text{ref}}_i\}_{i=1}^{N}$. \review{It is important to note that the classification of these instances is solely determined by the first stage. The predicted category labels $\{c_i\}_{i=1}^{N}$ are associated with their corresponding masks $\mathcal{M}^{\text{ref}}_{i}$ via index matching, as the point prompts and labels are generated and processed in a consistent order. The SAM adaptation stage focuses exclusively on binary mask generation.}

\subsection{Automatic Point Prompt Generation}
\label{method b}
To automate nuclear instance segmentation using SAM, the model requires one prompt per nucleus instance in a given pathological image. Conventional approaches often rely on bounding boxes to delineate each object before segmentation. However, in densely packed and overlapping nuclear regions, bounding boxes tend to collide and incorporate excessive background noise, which adversely affects segmentation accuracy. To overcome this limitation, we employ precise point prompts that directly indicate nuclear centroids, enabling more accurate and robust instance segmentation.

\subsubsection{Anchor Point Learning}
Inspired by \citep{song2021rethinking}, we introduce a lightweight prompt learner. As illustrated in Fig. \ref{fig:pipeline}, our model operates according to the following pipeline. 
First, the input image $I$ is processed by an image encoder $E_p$ for feature extraction. We adopt a Feature Pyramid Network (FPN) with a ConvNeXt-S backbone to produce multi-scale feature maps $\{F_{l}^{p}\}_{l=2}^{L_p}$. Here, each feature map $F_l^p$ has a spatial resolution of $(H/{2^l}, W/{2^l})$, and $L_p$ denotes the depth of the encoder. 
Second, we initialize a set of learnable anchor points $A = \{a_i\}_{i=1}^{M}$ by uniformly distributing them over the spatial dimensions of the input image $I$, where $M$ is the number of anchors. These anchor points are first refined by an offset network, which predicts an initial adjustment to obtain updated positions $\{\hat{a}_i\}_{i=1}^{M}$.
Third, for each anchor point $\hat{a}_i$, we sample its corresponding features from all scale levels of the feature maps $F_l^{p}$ via bilinear interpolation. The sampled multi-scale features are then concatenated into a unified descriptor. 
This unified descriptor is fed into two task-specific heads: a regression head $H_{reg}$ and a classification head $H_{cls}$. The regression head $H_{reg}$, implemented as a two-layer MLP with ReLU and Dropout, predicts an offset $\Delta_i$ to further refine the anchor $\hat{a}_i$ towards a ground-truth nucleus centroid, yielding the final anchor position ($\tilde{a}_i = \hat{a}_i + \Delta_i$). The classification head $H_{cls}$, also a two-layer MLP, predicts the category logits $q_i \in \mathbb{R}^{C + 1}$ for the anchor. 

To enhance the model's awareness of the foreground, we incorporate an auxiliary segmentation task. We utilize the feature map $F_2^{p}$ for this task due to its high spatial resolution, which is beneficial for capturing detailed foreground structures. The feature map is processed by an auxiliary segmentation head $H_{seg}$ (composed of two convolutional layers with batch normalization and ReLU) to produce a binary foreground mask.

\subsubsection{Anchor Point Matching}
During the training stage, we employ the Hungarian algorithm \citep{kuhn1955hungarian, song2021rethinking} to match the predicted anchor points with ground-truth annotations to define the training objectives. Specifically, we formulate the matching between the $M$ anchors and the $G$ ground-truth nuclear centroids as a bipartite matching problem, where $G$ denotes the total number of nuclei instances in the ground-truth.

During the bipartite matching, the matching score $w_{i,j}$ is defined to quantify the suitability of assigning the $i$-th anchor to the $j$-th ground-truth centroid. It is defined as a combination of classification confidence and spatial proximity:
\begin{equation}
{w}_{i,j} = \sigma (q_i)[c_j] - \alpha\| \tilde{a}_i - pos_j \|_2, \\
\end{equation}
where $c_j$ denotes the class label of the $j$-th nucleus, $pos_j$ represents its position, $\sigma (q_i)[c_j]$ is the corresponding class probability from the $i$-th anchor, $\sigma(\cdot)$ represents the function of softmax, and $\alpha$ is a balancing weight. The $l_2$ distance $\| \tilde{a}_i - pos_j \|_2$ measures the spatial discrepancy. A higher score $w_{i,j}$ indicates a more favorable match, as it implies both high classification confidence for the correct class and close spatial proximity to the ground-truth.
The Hungarian algorithm efficiently finds the matching that maximizes the total score across all assigned pairs. Anchors successfully matched to a ground-truth centroid are selected as positive samples, while unmatched anchors are assigned to the background class. 

After matching, we obtain $\mathcal{I}_{pos}$ and $\mathcal{I}_{neg}$ which denote the index sets of matched (positive) and unmatched (negative) anchors, respectively. Note that the number of positive anchors $|\mathcal{I}_{pos}|$ equals the number of ground-truth centroids $G$, and $|\mathcal{I}_{pos}| + |\mathcal{I}_{neg}| = M$, where $M$ is the total number of anchor points.

The total training loss for the prompt learner integrates the objectives of classification, regression, and auxiliary segmentation:
\begin{equation}
\begin{aligned}
\mathcal{L}_{total}^p &= \mathcal{L}_{cls} + \mathcal{L}_{reg} + \mathcal{L}_{seg}, \\
\mathcal{L}_{cls} &= -\frac{1}{M} ( \sum_{i\in \mathcal{I}_{pos}} \log \ \sigma (q_i)[c_i] + \beta \sum_{z \in \mathcal{I}_{neg}} \log \ \sigma (q_z)[\varnothing] ),\\
\mathcal{L}_{reg} &= \frac{\gamma}{G}( \sum_{i\in \mathcal{I}_{pos}} \| \tilde{a}_i - pos_i \|_2),
\end{aligned}
\end{equation}
where $\varnothing$ indicates the background class.  $\mathcal{L}_{seg}$ is the Focal Loss \citep{lin2017focal} between the predicted probability map and the ground-truth binary mask. The hyperparameters $\beta$ and $\gamma$ balance the contributions of the negative samples and the regression task, respectively. Note that for each positive anchor $i \in \mathcal{I}_{pos}$, $c_i$ and $pos_i$ denote the class and position of its specifically matched ground-truth centroid, respectively.

\subsubsection{Point Prompt Selection}
During inference, the trained model generates point prompts by filtering the refined anchor points $\{\tilde{a}_i\}_{i=1}^M$ based on their classification results. Specifically, anchors predicted as the background class are discarded. The remaining $N$ foreground anchors constitute the final prompt set $\mathcal{P} = \{p_i\}_{i=1}^N$. \review{These point prompts, along with their corresponding predicted category labels $\mathcal{C} = \{c_i\}_{i=1}^{N}$, are retained as an ordered set. While only the point prompts $\mathcal{P}$ are fed into the subsequent SAM adaptation stage to generate masks, the category labels $\mathcal{C}$ are preserved to be later associated with the output masks, ensuring that each segmented instance receives its predicted class.}

\begin{algorithm}[t]
    \caption{Prompt Learning Stage of CFR-SAM}
    \label{alg:prompt_learning}
    \begin{algorithmic}[1]
        \INPUT Image $I \in \mathbb{R}^{3 \times H \times W}$; during training, ground-truth nuclear centroids with class labels.
        \OUTPUT Point prompts $\mathcal{P} = \{p_i\}_{i=1}^{N}$ and corresponding category labels $\mathcal{C} = \{c_i\}_{i=1}^{N}$ (for inference).
        
        \STATE Extract multi-scale feature maps $\{F_l^p\}_{l=2}^{L_p}$ using encoder $E_p$.
        \STATE Feed $F_2^p$ into an auxiliary segmentation head $H_{seg}$ to predict foreground mask.
        \STATE Initialize $M$ anchor points $A = \{a_i\}_{i=1}^M$ uniformly over image space and apply preliminary coordinate adjustment to obtain $\hat{A} = \{\hat{a}_i\}_{i=1}^M$.
        
        \FOR{each anchor $\hat{a}_i$}
            \STATE Sample multi-scale features from $\{F_l^p\}_{l=2}^{L_p}$ and concatenate into a unified descriptor.
            \STATE From this descriptor, predict offset $\Delta_i$ via regression head $H_{reg}$ (yielding $\tilde{a}_i = \hat{a}_i + \Delta_i$) and category logits $q_i$ via classification head $H_{cls}$.
        \ENDFOR

        \IF {training}
            \STATE Match $\{\tilde{a}_i\}_{i=1}^M$ to ground-truth nuclear centroids via Hungarian algorithm (Eq.(1)), based on classification confidence and spatial proximity.
            \STATE Compute total loss $\mathcal{L}_{total}^{p}$ (Eq.(2)), and update network parameters.
        \ELSE[inference]
            \STATE Filter anchors by discarding those predicted as background, retain remaining $N$ anchors as point prompts $\mathcal{P} = \{p_i\}_{i=1}^N$ with their predicted category labels $\mathcal{C} = \{c_i\}_{i=1}^N$.
            \RETURN $\mathcal{P}, \mathcal{C}$
        \ENDIF
        
    \end{algorithmic}
\end{algorithm}

\input{Module1}

\subsection{Multi-Scale Adaptive Local-Aware Adapter}
\label{method c}

While SAM's rich prior and powerful global context modeling have proven effective across numerous tasks, directly applying SAM to nucleus segmentation via full fine-tuning is hindered by its substantial computational cost and inherent limitation in capturing fine-grained local information. 

To address these limitations, we propose MALA, a parameter-efficient fine-tuning method. By introducing minimal parameters into the frozen Transformer backbone, MALA bridges the domain gap to adapt SAM for nucleus segmentation. Unlike previous PEFT methods that primarily rely on linear transformations along feature channels, MALA dynamically generates pixel-wise, multi-scale convolutional kernels conditioned on the input features. This allows the model to adaptively assign the optimal receptive field for different nucleus regions within a low-rank space, thereby extracting region-specific, fine-grained features crucial for accurate nuclei segmentation.

As illustrated in Fig. \ref{fig:module1}, the trainable MALA module is inserted into each frozen Transformer block. This design enhances the model's ability to capture intricate nuclear structures while preserving its pre-trained capacity for global feature modeling.

\subsubsection{The Detailed Architecture of MALA}
\review{The architecture of MALA is illustrated in Fig.\ref{fig:module1}. Given an input feature map $X \in \mathbb{R}^{C \times H^{'} \times W^{'}}$, we first apply layer normalization and a residual connection, followed by a down-projection layer that compresses the channel dimension from $C$ to $\hat{C}$ to reduce computational cost, where $C \gg \hat{C}$. The resulting feature $X^{'}$ then enters the core adaptive multi-scale convolution module.}

To handle substantial variations in nuclear scale and density across different images, we introduce an adaptive multi-scale convolution module. This module dynamically generates pixel-wise kernels conditioned on $X^{'}$, allowing it to assign an optimal receptive field to each location. Specifically, for each candidate kernel size $k \in K = \{3,5,7\}$, we employ a dedicated $3\times3$ convolution layer to produce the depth-wise convolution weights for that scale:

\begin{equation}
W_{conv}^{k} = Conv_{3 \times 3}^{k}(X^{'}), \quad W_{conv}^{k} \in \mathbb{R}^{k^2 \times H^{'} \times W^{'}}.
\end{equation}
Here, $Conv_{3 \times 3}^{k}$ denotes a separate convolutional layer for each scale $k$, with an output channel dimension of $k^2$.

Simultaneously, the unfold operation extracts image patches of spatial size $k \times k$ from $X^{'}$:
\begin{equation}
X_{unf}^{k} = unfold(X^{'}, k), \quad X_{unf}^{k} \in \mathbb{R}^{\hat{C} \times k^2 \times H^{'} \times W^{'}}.
\end{equation}

The generated kernel $W_{conv}^{k}$ then performs a depth-wise convolution on its corresponding patches $X_{unf}^{k}$. The outputs from all scales are fused by averaging and combined with the original input via a residual connection:
\begin{equation}
\bar{X}^{'} = \frac{1}{|K|}\sum_{k\in K}(W_{conv}^{k} \ast X_{unf}^{k}) + X^{'}, 
\end{equation}
where $\ast$ denotes the depth-wise convolution operation.

While depth-wise convolution is efficient, it lacks inter-channel communication. \review{To compensate, a $1\times1$ convolution is applied to $\bar{X}^{'}$ to enable cross-channel communication, followed by a residual connection. }

\review{As the final step, a GeLU activation function is applied, followed by an up-projection layer with parameters $W_{up} \in \mathbb{R}^{\hat{C} \times C}$ to restore the channel dimension, and another residual connection is added to produce the output.}

\input{Module2}

\subsection{Hierarchical Modulated Fusion Module}
\label{method d}
SAM's encoder captures information at markedly different levels of granularity across layers: shallow layers focus on low-level details such as edges and textures, whereas deep layers encode high-level semantics. Previous methods \citep{chen2024sam,ma2024segment} often rely solely on deep features, neglecting the crucial role that shallow-layer details play in precisely localizing nuclei within complex histological sections. However, simply summing or concatenating these multi-level features is suboptimal due to their semantic misalignment. 
To overcome this limitation, we introduce the HMFM module, which adaptively modulates features from different encoder layers, thereby ensuring the effective integration of fine-grained details with high-level semantics. This integration significantly improves the accuracy of both nuclear identification and segmentation.

As illustrated in Fig.\ref{fig:module2}, our module leverages a channel-wise attention unit to modulate encoder features from multiple SAM layers, which facilitates cross-layer representation integration. Concretely, we select the output features $X_l$ from layer set $ S = \{\frac{1}{4}L, \frac{1}{2}L, \frac{3}{4}L, L\}$, where $L$ denotes the layer number of SAM encoder. For each $X_l$ we dynamically generate channel-wise weights $W_l^{m}$ and apply them to harmonize the inter-layer representations as follows:

\begin{equation}
W_l^{m} = Sigmoid(GAP(Conv_{1 \times 1}(X_l))), \\
\end{equation}
where GAP is the Global Average Pooling. The modulated feature $X_i^{m}$ is obtained by a residual gating mechanism:

\begin{equation}
{X_l^{m}} = W_l^{m} \odot X_l + X_l, \\
\end{equation}
where $\odot$ denotes element-wise multiplication with broadcasting. Finally, we aggregate the modulated features into the final representation $X_f$:
\begin{equation}
\begin{aligned}
X_f &= \sum_{l \in S}X_l^m.
\end{aligned}
\end{equation}

$X_f$ then serves as the image embedding for the mask decoder, seamlessly integrating fine-grained structural details with high-level semantic context.

\input{Module3}

\begin{algorithm}[t]
    \caption{SAM Adaptation Stage of CFR-SAM}
    \label{alg:sam_adaptation}
    \begin{algorithmic}[1]
        \INPUT Input image $I$ and point prompts $\mathcal{P} = \{p_i\}_{i=1}^{N}$.
        \OUTPUT For each nucleus $i$: initial mask $\mathcal{M}^{ori}_i$, refined mask $\mathcal{M}^{ref}_i$ and boundary mask $\mathcal{M}^{bound}_i$.
        
        \STATE Pass $I$ through SAM encoder with MALA (Sec. \ref{method c}), store intermediate features $X_l$ for $l \in S = \{\frac{1}{4}L, \frac{1}{2}L, \frac{3}{4}L, L\}$, where $L$ is the total number of encoder layers.
        
        \STATE Fuse $\{X_l\}_{l \in S}$ using HMFM (Sec. \ref{method d}) to obtain multi-level feature $X_f$.
        
        \STATE Encode $\mathcal{P}$ via prompt encoder to obtain embeddings $E_p$.
        
        \STATE Decode $X_f$ and $E_p$ to get initial masks $\mathcal{M}^{ori}$ and mask features $X_{mask}$.
        
        \STATE Upsample $X_f$ to match $X_{mask}$ resolution, fuse them, and upsample to full resolution to obtain $X_{mask}^{up}$.
        
        \STATE Extract $X_{mc}$ from $I$ via MCBE (Sec. \ref{MCBE}) to capture multi-scale edge details.
        
        \STATE Fuse $X_{mask}^{up}$ and $X_{mc}$ via BGFF (Sec. \ref{BGFF}) to produce boundary-oriented feature $X_{bound}$ and predict boundary mask $\mathcal{M}^{bound}$.
        
        \STATE Refine $X_{mask}^{up}$ with $X_{bound}$ via residual bottleneck to obtain enhanced features $X_{mask}^{ref}$.
        
        \STATE Generate refined masks $\mathcal{M}^{ref}$ from $X_{mask}^{ref}$ using mask head $H_{mask}$.
        
        \STATE Compute $\mathcal{L}_{total}$ (Eq. (15)-(16)) supervising $\mathcal{M}^{ori}$, $\mathcal{M}^{ref}$, and $\mathcal{M}^{bound}$.
        
        \RETURN $\{(\mathcal{M}^{ori}_i, \mathcal{M}^{ref}_i, \mathcal{M}^{bound}_i)\}_{i=1}^{N}$
    \end{algorithmic}
\end{algorithm}

\subsection{Boundary-Guided Mask Refinement}
\label{method e}
The feature $X_f$ produced by the preceding modules fuses rich multi-scale local details with global semantics. However, the mask decoder of SAM employs a prediction paradigm that generates low-resolution masks followed by upsampling. This inherent design often leads to a loss of spatial precision and blurred object boundaries in the final segmentation.

To address this issue, we introduce the BGMR which explicitly amplifies boundary-related signals in the feature space, thereby refining the coarse masks produced by the decoder to achieve sharper delineation of individual nuclei.

\subsubsection{Initial Coarse Mask Generation}
As illustrated in Fig.\ref{fig:pipeline}, we first obtain coarse mask features $X_{mask}$. The previously derived point prompts $\mathcal{P} = \{p_i | (p_i,c_i) \in \mathcal{D}\}$ are encoded into prompt embeddings by the prompt encoder and then interact with the image feature $X_f$ in the mask decoder, producing nucleus-specific representations $X_{mask}$ and initial masks $\mathcal{M}^{ori}$ for each prompt.

However, these initial masks are typically coarse and lack fine boundary details due to the low-resolution prediction paradigm of the decoder. As illustrated in Fig.\ref{fig:module3}, to refine the spatial precision of the final masks, we first align and fuse the semantic feature $X_{f}$ with the coarse mask feature $X_{mask}$. Since $X_f$ is at $1/16$ of the input resolution and $X_{mask}$ is at $1/4$, we upsample $X_f$ to the $1/4$ resolution using bilinear upsampling followed by convolutions. \review{Specifically, we apply two rounds of $2\times$ upsampling, each followed by a $3\times3$ convolution, to obtain $X_f^{up}$. Then, $X_f^{up}$ and $X_{mask}$ are concatenated along the channel dimension and fused via a $1\times1$ convolution, yielding the fused feature $\hat{X}_{mask}$. Finally, $\hat{X}_{mask}$ is further upsampled to the original input resolution, again using two $2\times$ upsampling steps with intermediate $3\times3$ convolutions, to produce the upsampled mask feature $X_{mask}^{up}$ for subsequent boundary-guided refinement.}

\subsubsection{Multi-Context Boundary Enhancement} 
\label{MCBE}
To fully preserve boundary details at the original resolution, we extract fine-grained structures directly from the input image. However, the extreme variability in nuclear size and density within whole slide images requires multi-scale contextual information for accurate boundary discrimination. To this end, we introduce the MCBE module.

\review{The MCBE module operates on the base feature map $X_{init}$, which is obtained by applying a $3\times3$ convolution to the input image $I$. From $X_{init}$, we first apply another $3\times3$ convolution to generate shallow features $X_{low}^0$ that contain high-frequency details.}

To capture complementary contextual information from fine to coarse levels, we employ a progressive smoothing strategy. By sequentially applying average pooling and convolution operations, we produce a multi-scale feature set $\{X_{{low}}^{1}, X_{{low}}^{2}, X_{{low}}^{3}\}$ at the original spatial resolution. Each successive feature layer integrates broader local contextual information than the last, as follows:

\begin{equation}
X_{low}^t = Conv_{3\times3}(AP(X_{low}^{t-1})), \quad \text{for } t \in \{1,2,3\}.
\end{equation}
Here, $AP$ denotes a 3$\times$3 average pooling layer with stride 1 and padding 1, which smooths the features and enables subsequent convolutions to process coarser structural information.

These smoothed features $\{X_{low}^{1}, X_{low}^{2}, X_{low}^{3} \}$ are then fed into Boundary Enhancement (BE) block. The BE block effectively amplifies boundary signals by highlighting the discrepancy between input feature and its locally smoothed version, defined as:
\begin{equation}
X_{be}^t = Conv_{1\times1}(X_{low}^t - AP(X_{low}^t)) + X_{low}^t.
\end{equation}

The set $\{X_{\text{low}}^0, X_{\text{be}}^1, X_{\text{be}}^2, X_{\text{be}}^3\}$ forms our multi-contextual representation for boundary perception. To preserve the original high-frequency details in $X_{\text{low}}^0$ that are crucial for sharp edges, we apply the BE only to the smoothed features $\{X_{\text{low}}^{t}\}_{t=1}^{3}$. This design ensures that the finest structures from the initial feature are retained.

\review{To obtain the final output $X_{mc}$, all features in the set are concatenated along the channel dimension, compressed via a $1\times1$ convolution, and integrated with the base feature $X_{init}$ through a residual connection.}

\subsubsection{Boundary-Guided Feature Fusion} 
\label{BGFF}
Due to the limited representation capacity of the lightweight detail extraction module, the extracted detail feature $X_{mc}$ lacks sufficient semantic context. Directly fusing it with the upsampled semantic feature $X_{mask}^{up}$ would lead to noise coupling, thereby compromising the original discriminative power of $X_{mask}^{up}$. 

To address this, we introduce an auxiliary boundary segmentation task to guide the fusion process itself. Through the boundary supervision signal, the model is compelled to learn to select and combine the information most relevant to boundaries during the fusion of $X_{mask}^{up}$ and $X_{mc}$, thereby learning a denoised, boundary-oriented fused feature $X_{bound}$. We refer to the module designed for this purpose as the BGFF. Its detailed structure is shown in Fig. \ref{fig:module3}.

\review{Specifically, to reduce computational cost, we first split $X_{mask}^{up}$ along the channel dimension into two low-dimensional groups, denoted as $X_{mask,1}^{up}$ and $X_{mask,2}^{up}$. Subsequently, each group is concatenated with the boundary-enhanced feature $X_{mc}$ and processed through two parallel fusion branches.} In one branch, we apply a standard convolution layer to fuse the features, while in the other one, we employ a deformable convolution layer to adaptively align and fuse the semantic and boundary information. The outputs of both branches are then modulated by a Spatial Attention (SA) module \citep{woo2018cbam} to achieve preliminary focus on important regions: 

\begin{equation}
Att_1 = \text{SA}(Conv([X_{mask,1}^{up}, X_{mc}])),\\
\end{equation}
\begin{equation}
Att_2 = \text{SA}(DConv([X_{mask,2}^{up}, X_{mc}])),\\
\end{equation}
\begin{equation}
\hat{X}_{mask,v}^{up} = Att_vX_{mask,v}^{up} + X_{mask,v}^{up}, \quad \text{for } v \in \{1,2\}.
\end{equation}

\review{The modulated features $\hat{X}_{mask,1}^{up}$ and $\hat{X}_{mask,2}^{up}$ are then concatenated and fused via a $1\times1$ convolution to obtain the initially fused feature $X_{bound}$. Subsequently, a boundary segmentation head $H_{bound}$, composed of convolution layers, is applied to $X_{bound}$ to predict the boundary map $\mathcal{M}_{bound}$, which provides the supervision signal to guide the fusion process.}

\subsubsection{Boundary-Guided Mask Refinement} 
After obtaining the boundary-oriented feature $X_{bound}$, we utilize it to reinforce the semantic feature. \review{First, $X_{bound}$ is transformed into spatial weights via sigmoid and multiplied with $X_{mask}^{up}$ to enhance the responses in boundary regions.} This weighted feature is then passed through a residual bottleneck layer (consisting of a $1\times1$ convolution, a deformable convolution, and another $1\times1$ convolution) and added back to $X_{mask}^{up}$, producing the refined feature $X_{mask}^{ref}$:

\begin{equation}
X_{mask}^{ref} = \text{Bottleneck}(Sigmoid({X}_{bound}) \odot X_{mask}^{up}) + X_{mask}^{up}.
\end{equation}

\review{Finally, a mask head $H_{mask}$ (composed of two $3\times3$ Conv-BN-ReLU blocks followed by a $1\times1$ convolution) generates the refined segmentation $\mathcal{M}^{ref}$ from $X_{mask}^{ref}$.}

\input{pannuke_instance}

\subsection{Training Objective}
To train our framework, we adopt a multi-task loss function that supervises both mask prediction and boundary awareness. The overall objective combines supervision for the original and refined masks from SAM, along with explicit boundary prediction:
\begin{equation}
\mathcal{L}_{total} = \mathcal{L}_{mask}(\mathcal{M}^{ori}, \mathcal{M}^{gt}) + \mathcal{L}_{mask}(\mathcal{M}^{ref}, \mathcal{M}^{gt}) + \omega\mathcal{L}_{bound},
\end{equation}
where $\mathcal{L}_{mask}$ denotes SAM mask loss, $\mathcal{M}^{ori}$ is the initial mask predicted by the mask decoder, $\mathcal{M}^{ref}$ is the refined mask produced by the BGMR module, $\mathcal{L}_{bound} = \mathcal{L}_{BCE}(\mathcal{M}_{bound}, \mathcal{M}_{bound}^{gt})$ is the boundary prediction loss, and $\omega$ balances their contributions.
The mask loss $\mathcal{L}_{mask}$ for a predicted mask $\mathcal{M}$ is composed of:
\begin{equation}
\mathcal{L}_{mask}(\mathcal{M}, \mathcal{M}^{gt}) = \lambda\mathcal{L}_{FL} + \mathcal{L}_{DL} + \mathcal{L}_{IoU},
\end{equation}
where $\mathcal{L}_{FL}$ and $\mathcal{L}_{DL}$ represent the focal and dice losses between $\mathcal{M}$ and the ground-truth $\mathcal{M}^{gt}$, respectively. $\mathcal{L}_{IoU} = \mathcal{L}_{MSE}(\hat{o}, o)$ penalizes the discrepancy between the predicted IoU $\hat{o}$ and ground-truth IoU $o$, and $\lambda$ is a weighting factor.

\section{Experiments}
\input{statistic}
\subsection{Datasets}
\textbf{PanNuke} \citep{gamper2019pannuke, gamper2020pannuke} is a benchmark for simultaneous nucleus instance segmentation and classification, notable for its diversity and scale. It contains 7,904 H$\&$E stained images ($256 \times 256$ pixels) from 19 tissue types, with 189,744 nuclei annotated across five categories. Following the standard evaluation protocol \citep{gamper2020pannuke}, we adopt a three-fold cross-validation strategy and report the average performance.

\textbf{CPM-17} \citep{vu2019methods} consists of 64 H$\&$E stained images with 7,570 annotated nuclei. The images have resolutions of either $500 \times 500$ or $600 \times 600$ pixels. The dataset is equally split between training and testing, with 32 images allocated to each set.

\textbf{MoNuSeg} \citep{kumar2019multi} provides annotations for nucleus segmentation but does not include nuclear classification. This multi-organ dataset contains tissue samples from breast, liver, kidney, prostate, bladder, colon, and stomach. It includes 44 H$\&$E stained images, each with a resolution of $1000 \times 1000$ pixels.

\input{pannuke_det_cls}
\subsection{Evaluation Metrics}
To thoroughly evaluate the performance of our method, we employ the following metrics. For instance segmentation, we use the Adjusted Jaccard Index (AJI), binary Panoptic Quality (bPQ), and multi-class Panoptic Quality (mPQ), which together provide a comprehensive assessment of object-level segmentation accuracy \citep{kumar2017dataset, graham2019hover, horst2024cellvit}. For classification, we employ Precision (P), Recall (R), and $F_1$-score ($F_1$) for each nucleus type. All metrics are computed for each image, and the overall results for each dataset are reported as averages across all images. 

\subsection{Implementation Details}
\label{Implementation}
\review{To evaluate the proposed CFR-SAM framework, we develop two variants based on different SAM backbones: Ours-H (utilizing SAM-Huge) and Ours-B (utilizing SAM-Base). Ours-H represents the full capacity of our approach and is designed to explore the upper bound of segmentation performance, serving as the primary variant for comparison with state-of-the-art methods across all datasets. Ours-B, with its parameter scale aligned with mainstream lightweight models, is introduced primarily for two purposes: (1) to serve as the backbone for extensive ablation studies (Sec. 4.5) due to its computational efficiency, and (2) to provide a fair basis for computational efficiency comparison (Sec. 4.6) against methods of similar model size.} 

We implement our model using PyTorch on a workstation equipped with a single NVIDIA RTX 3090 GPU. Following common practices, we use both SAM-Base and SAM-Huge as backbones and optimize the model with Adam, using an initial learning rate of $1 \times 10^{-4}$. Our training pipeline is divided into two consecutive stages with distinct configurations.

\textbf{Stage 1: Prompt Learning.}
In this stage, the distance coefficient $\alpha$ and the regression loss weight $\gamma$ are set to 0.1 and $5 \times 10^{-3}$ respectively across all datasets. The other hyperparameters are tailored to each dataset's characteristics. For PanNuke, we use an FPN length $L_p$ of 3, a classification loss coefficient $\beta$ of 0.4, a batch size of 16, and train for 200 epochs. For CPM-17, the configuration uses $L_p=4$, $\beta=0.25$, a batch size of 8, and 200 epochs. For the smaller MoNuSeg dataset, we set $L_p=4$, $\beta=0.2$, a batch size of 4, and extend the training to 600 epochs to ensure sufficient learning.

\textbf{Stage 2: SAM Adaptation.}
During the second stage, a unified set of hyperparameters is applied across all datasets for consistency: the boundary loss weight $\omega$ is set to $1 \times 10^{-2}$, the overall loss weight $\lambda$ is 20, the batch size is 2, and the training runs for 200 epochs. To ensure training efficiency and manage GPU memory constraints, we randomly sample a maximum of 25 nuclei from each image during training. For data augmentation, we adopt the same strategy as CellViT \citep{horst2024cellvit}.

\review{We adopt this two-stage strategy due to the non-differentiable nature of the point prompt selection process. The prompt learner generates candidate points that are filtered based on classification confidence before being passed to SAM's mask decoder. This discrete selection operation blocks gradient flow from the segmentation loss back to the prompt learner, making end-to-end joint optimization unstable, especially during early training when the prompt learner is not yet mature and the selected points are often noisy or empty, causing SAM to receive invalid inputs. In contrast, the decoupled design first trains the prompt learner independently, then fine-tunes SAM with stable prompts, ensuring training stability.}

\review{For datasets without official validation splits (CPM-17, MoNuSeg), we randomly hold out 15\% of training images as a validation set for model selection and early stopping. Test sets are never used during training or tuning. For PanNuke, we follow the standard three-fold protocol. All hyperparameters are selected based on validation performance.}

\input{pannuke_mpq}
\input{h_compare}
\input{instance_cls}
\subsection{Comparison With SOTA Methods}
\label{sota_comparison}
In this section, we evaluate the proposed method against state-of-the-art approaches on three challenging benchmarks: PanNuke, CPM-17, and MoNuSeg.

\subsubsection{PanNuke Dataset}
As summarized in Table \ref{tab:pannuke_instance}, our method achieves substantial improvements over existing competitors in both bPQ and mPQ. Specifically, Ours-B surpasses the previous best model by 1.30\% in bPQ and 1.11\% in mPQ, while Ours-H further elevates these gains to 1.63\% in bPQ and 1.35\% in mPQ. \review{The consistent performance improvement from Ours-B to Ours-H demonstrates the scalability of our framework: leveraging a larger SAM backbone yields proportional gains in segmentation accuracy. Notably, despite having significantly fewer parameters than fully fine-tuned models, Ours-B already achieves state-of-the-art performance, confirming the effectiveness of our parameter-efficient design even with a lightweight backbone.} 

\review{To assess the robustness of our method, we conduct five independent runs on the PanNuke dataset (each with three-fold cross-validation), resulting in 15 evaluation results per model. The mean and standard deviation of the key metrics are reported in Table \ref{tab:statistic}. The low standard deviations, for instance below 0.20\% for bPQ and below 0.13\% for mPQ across both model variants, demonstrate that our method consistently achieves strong performance with minimal variability.}

\review{Furthermore, to evaluate the overall instance-level segmentation quality and address potential concerns regarding over-segmentation, we examine the AJI, which is sensitive to both over- and under-segmentation by jointly assessing segmentation accuracy and instance assignment correctness. As reported in Table \ref{tab:statistic}, our method achieves substantial and consistent AJI improvements over the previous state-of-the-art (CellViT-H, 0.6519). Specifically, Ours-H attains an AJI of 0.6887 $\pm$ 0.0004, with similarly low variance, indicating that the observed PQ gains are accompanied by more accurate instance identification and counting, not merely finer boundary predictions. This improvement further validates that our framework mitigates over-segmentation risks through its centroid-based prompting and boundary-guided refinement.}

Notably, compared to the fully fine-tuned (FFT) SOTA approach, our method achieves superior performance while training only 10.6\% of the parameters (74.2M vs. 699.7M), demonstrating our method's efficiency. It is worth noting that the trainable parameters of Ours-H are distributed as 55.1M in the first stage and 19.1M in the second.

In terms of nucleus classification, as shown in Table \ref{tab:pannuke_detect}, our method attains the highest overall detection $F_1$-score of 83\%. The category-specific $F_1$-scores for the five cell types are: Neoplastic (76\%), Epithelial (78\%), Inflammatory (70\%), Connective (62\%), and Dead (45\%). \review{It is worth noting that these classification metrics are produced entirely by the first stage, with the predicted category labels subsequently associated with their corresponding masks through point prompt matching.}

Furthermore, Table \ref{tab:pannuke_mpq} details the segmentation performance per category, where our method achieves the highest PQ scores in three key tissue types: Neoplastic (0.599), Epithelial (0.589), and Connective (0.433), demonstrating its consistent superiority across diverse nuclear morphologies.

\subsubsection{CPM-17 and MoNuSeg Datasets}
To assess generalization capability, we further evaluate our method on the CPM-17 and MoNuSeg datasets. As shown in Table \ref{tab:cpm17}, on CPM-17, our approach outperforms the leading prior method by 0.4 points in AJI and 1.3 points in PQ, achieving scores of 72.5\% and 71.9\%, respectively. Similarly, on MoNuSeg (Table \ref{tab:monuseg}), we observe improvements of 0.7 points in AJI and 2.8 points in PQ, resulting in final scores of 66.8\% and 66.2\%. These consistent gains across multiple benchmarks strongly validate the robustness and general applicability of our method in varied nucleus instance segmentation scenarios.

\subsubsection{Qualitative Results}
As shown in Figs. \ref{fig:h_compare} and \ref{fig:instance_cls_compare}, we present visual comparisons between our method and CellViT-H for nucleus instance segmentation and classification, respectively. The results demonstrate that our model produces smoother and more accurate nuclear boundaries, particularly in regions with overlapping nuclei. These improvements can be attributed to our model’s enhanced ability to capture fine-grained local details and boundary information.

\input{cpm17}
\input{monuseg}

\subsection{Ablation Study}
To evaluate the effectiveness of our proposed modules, we conduct a series of ablation studies. For efficiency, all ablation experiments are performed using a fixed data split (trained on Fold 1, validated on Fold 2, and tested on Fold 3) of the PanNuke dataset, with the pre-trained SAM-B as the backbone. \review{We choose the lightweight Ours-B variant for ablation studies because it allows efficient module-wise analysis while maintaining architectural consistency with Ours-H, ensuring that insights gained are transferable to the larger model.}

\subsubsection{Analysis on Different Fine-Tuning Method}
To enable a comprehensive comparison with our approach, we employ two fine-tuning methods, FFT and freezing the entire backbone while training only the head. The results are presented on the first three rows of Table \ref{tab:ablation}. Our approach outperforms these two methods with only a small number of trainable parameters, demonstrating its efficiency and effectiveness. \review{As shown in Figs. \ref{fig:segmentation_comparison}, we present additional visual comparisons with fully fine-tuned SAM, where our method demonstrates superior performance in handling clustered nuclei and boundary details.}

\subsubsection{Analysis on Multi-Scale Adaptive Local-Aware Adapter}
\review{As shown in the MALA section of Table \ref{tab:ablation}, removing MALA from the encoder leads to a drop of 0.66\% in mPQ and 0.88\% in bPQ, demonstrating its contribution to fine-grained feature adaptation.} To validate the effectiveness of the learnable kernel, we compare adaptive convolutional kernels with static kernels. The results indicate that adaptive kernels provide a 0.13\% and 0.17\% performance gain for bPQ and mPQ, respectively, demonstrating their ability to moderately enhance model robustness and flexibly adapt to different complex scenarios. \review{Moreover, we include a higher parameter static kernel variant which still underperforms the learnable version, further indicating that adaptivity offers advantages beyond simply adding parameters.}

\input{ablation}

\subsubsection{Analysis on HMFM}
We evaluate the impact of HMFM through ablations in Table \ref{tab:ablation}. First, removing HMFM altogether results in performance degradation, with mPQ and bPQ dropping by 0.09\% and 0.13\%, respectively, which establishes the necessity of a dedicated fusion module. To further validate the superiority of HMFM's adaptive modulation, we compare it against three standard fusion strategies: addition, averaging, and concatenation. Notably, the proposed HMFM consistently achieves the best performance on both metrics, outperforming all the alternative fusion strategies. This consistently demonstrates the effectiveness of adaptively modulating features from different stages prior to fusion.

\subsubsection{Analysis on BGMR}
As shown in the BGMR section of Table \ref{tab:ablation}, without employing BGMR to refine coarse mask, directly using upsampled low-resolution mask as final predicted mask results in a decline in mPQ and bPQ by 0.56\% and 0.83\%, respectively. The visual comparisons in Fig. \ref{fig:ablation_refine} further demonstrate the effectiveness of BGMR in producing superior mask boundaries. \review{To control for the influence of increased parameters, we replace BGMR with a linear layer of equivalent size. As shown, our proposed module consistently achieves superior performance across the evaluation metrics.}

To validate the effectiveness of key components within BGMR, we conduct ablation studies on MCBE, BGFF, and the explicit boundary supervision. As shown in Table \ref{tab:ablation}, removing the MCBE module results in a 0.09\% and 0.12\% performance drop, as features extracted from the image lacked boundary enhancement before being fused with semantic features. This confirms the importance of explicitly reinforcing boundary information in features.

When the BGFF module is ablated, the boundary-enhanced features are directly used to guide semantic refinement and boundary prediction without prior feature alignment, weakening the representational capacity of the semantic features and leading to a 0.14\% and 0.22\% decrease in performance.

Finally, removing the explicit boundary supervision weaken the model’s ability to perceive boundary structures, resulting in a further performance degradation of 0.13\% and 0.21\%. These results collectively demonstrate the critical role of each component in improving boundary awareness and overall segmentation accuracy.

\input{fft_paper}
\input{refine}
\input{peft}
\input{ablation_peft}

\input{ablation_dimention}

\subsubsection{Analysis on Different PEFT Strategy}
To evaluate the effectiveness of our proposed MALA, we compare it against established PEFT methods, including Adapter \citep{houlsby2019parameter} and LoRA \citep{hu2022lora}, integrated into the image encoder of SAM for nuclei instance segmentation. To ensure a fair comparison, the low-rank dimensions of Adapter and LoRA are set to be identical to those used in MALA. As presented in Table \ref{tab:ablation_peft}, MALA demonstrates superior performance over these baselines. This result validates the importance of incorporating local detail information for accurate nuclei segmentation. \review{To further control for parameter count, we increase the intermediate dimension of Adapter to roughly match MALA's parameter count. Under this setting, increasing model capacity does not yield performance gains; in fact, it leads to slightly worse results.} Furthermore, visual comparisons in Fig. \ref{fig:ablation_peft} illustrate that MALA enables the model to capture fine-grained local details, allowing it to accurately segment nuclei of varying sizes and quantities, even in complex scenarios.

\subsubsection{Analysis on Different Low-Rank Dimension}
In parameter-efficient fine-tuning methods, the low-rank dimension is a critical hyperparameter. To investigate its impact, we conduct experiments with low-rank dimensions set to 16, 32, 64, and 128, respectively. As shown in Table \ref{tab:ablation_dimention}, the model achieves optimal performance when the low-rank dimension is set to 64. We argue that an excessively low rank (e.g., 16 or 32) provides too few trainable parameters, hindering sufficient knowledge transfer from the pre-trained model to the target domain. Conversely, an overly high dimension (e.g., 128) introduces an excess of trainable parameters, which may dilute the valuable pre-trained knowledge.

\input{mask_num}

\subsubsection{Analysis on Number of Sampled Nuclei}
\label{sec:ablation_sampling}
\review{During training, we randomly sample a maximum of 25 nuclei per image to balance computational efficiency and model performance. To verify that this choice does not introduce bias or affect robustness, we conduct an ablation study by varying the maximum number of sampled nuclei to 20, 25, 30, and 35, while keeping all other settings unchanged. The results are reported in Table \ref{tab:mask_num}.}

\review{As shown, the performance remains stable across all tested values, with negligible fluctuations (within 0.03\% for mPQ and 0.06\% for bPQ). This indicates that sampling 25 nuclei does not introduce noticeable bias and that the model's performance is robust to this hyperparameter choice. Based on these observations, we select 25 as the default value to balance training efficiency and GPU memory constraints, as using fewer than 25 would not significantly reduce computation in practice (the maximum is rarely reached in most images), while using more than 25 offers no performance gain but increases memory consumption.}

\input{box}
\subsubsection{Analysis on Prompt Type}
\review{To empirically validate the effectiveness of point prompts over bounding box prompts within our framework, we conduct an additional experiment by replacing point prompts with bounding boxes while keeping all other components unchanged. Specifically, we modify the Prompt Learner to generate bounding boxes instead of centroid points. All other modules, including MALA, HMFM, and BGMR, remain identical. The comparative results, shown in Table \ref{tab:box}, demonstrate that point prompts consistently and significantly outperform bounding box prompts, achieving improvements of approximately 2.4\% in mPQ and 3.0\% in bPQ. These results confirm that point prompts, by directly indicating nuclear centroids, offer more precise localization and cleaner contextual information, thereby enabling superior instance segmentation.}

\input{efficient_compare}

\subsection{Model Efficiency}
\review{As elaborated in Sec.~\ref{Implementation}, we introduce two variants of our method: Ours-H aims to explore the performance upper bound, while Ours-B is designed with a parameter scale comparable to most mainstream methods, serving as the primary basis for a fair efficiency comparison.} In terms of efficiency analysis, as illustrated in Table \ref{tab:efficient}, our proposed method demonstrates notable advantages in parameter efficiency and computational complexity. Compared to existing approaches that rely on heavyweight U-Net architectures and complex post-processing pipelines, our CFR-SAM framework employs a parameter-efficient fine-tuning strategy by keeping the SAM backbone frozen while introducing only a minimal number of trainable parameters. This design significantly reduces both model storage requirements and computational overhead in terms of MACs. It is worth noting that among all trainable parameters, the majority are concentrated in the first-stage prompt generation. The second-stage fine-tuning of SAM, in comparison, is very lightweight, requiring only 8.3M additional parameters in the base configuration (Ours-B) \review{and 19.1M in the huge configuration (Ours-H)}. This parameter-efficient architecture achieves an excellent balance between performance and computational cost. 

\review{It is important to clarify that our primary focus in this work is parameter-efficient adaptation, i.e., achieving strong performance while minimizing the number of trainable parameters during fine-tuning. Consequently, FPS is not our main optimization target. The FPS values reported in Table \ref{tab:efficient} are therefore intended for reference only. The lower throughput relative to some other methods stems from an intentional design trade-off:} to enhance local feature extraction, we introduce the MALA module, which employs multi-scale convolutional operations. While this design effectively captures fine-grained details, it inevitably adds computational overhead during inference. We accept this moderate reduction in speed in exchange for improved segmentation accuracy, a trade-off that we consider justified for precision-critical applications such as nuclei instance segmentation.

\section{Discussion and Limitation}
The core objective of this study is to address the key bottlenecks in applying SAM to nuclear instance segmentation: the high computational cost of full fine-tuning and the Transformer architecture's insufficient perception of local details. To eliminate complex post-processing pipelines, we choose to directly utilize SAM's mask decoder for instance segmentation. However, this choice introduce two new challenges: the underutilization of shallow encoder features and the boundary ambiguity caused by its upsampling predictions. The proposed CFR-SAM framework successfully addresses all the above challenges through a parameter-efficient strategy. Experimental results demonstrate that this method achieves high accuracy while significantly improving efficiency, offering a new perspective for leveraging large foundation models to solve specific medical imaging tasks.

\subsection{Local Context Enhancement: Bridging Global Priors and Local Precision}
While SAM’s global representation capability, learned from natural images, provides strong generalization, its effectiveness in nuclear segmentation critically depends on the perception of fine-grained local structures such as boundaries and morphology. A core innovation of CFR-SAM lies in its MALA module, which efficiently injects local perception into the frozen SAM backbone under the constraints of Parameter-Efficient Fine-Tuning. MALA is not merely a simple introduction of convolutions. Its key strength is the adaptive multi-scale mechanism within a low-rank space, which allows the model to dynamically adjust the receptive field based on local image content. The consistent performance gains from learnable convolutional kernels in our ablation studies (Table \ref{tab:ablation}) and the visualization comparison with othe PEFT methods (Fig. \ref{fig:ablation_peft}) demonstrate that this adaptive capability is crucial for handling the vast differences in nucleus size, density, and contrast, effectively compensating for the shortcomings of pure Transformer architectures in fine-grained medical image segmentation.

\subsection{Targeted Compensation Mechanisms for Decoder Deficiencies}
To fully leverage the potential of the SAM mask decoder for instance segmentation, it is essential to compensate for its two inherent architectural weaknesses. The proposed HMFM and BGMR are precisely targeted compensation solutions. HMFM intelligently fuses multi-level encoder features through an innovative modulation mechanism, ensuring that shallow information rich in spatial details is effectively utilized by the decoder. BGMR goes a step further by introducing explicit boundary supervision and a dedicated refinement process involving multi-context boundary enhancement and feature alignment, deeply integrating boundary awareness into the training process. Visual comparisons in Fig. \ref{fig:ablation_refine} confirm that BGMR is key to the model's ability to directly output high-quality instance masks.

\subsection{Parameter Efficiency and Generalization Capability}
CFR-SAM achieves performance comparable to or surpassing state-of-the-art methods on multiple challenging datasets (Tables \ref{tab:pannuke_instance}, \ref{tab:cpm17}, \ref{tab:monuseg}), demonstrating its strong generalization capability. More importantly, this performance improvement is achieved with a minimal parameter increase (as shown in Table \ref{tab:efficient}), highlighting the significant advantage of its parameter-efficient fine-tuning strategy. This indicates that through carefully designed adapter modules, the capabilities of large foundation models can be effectively transferred to the data-scarce medical imaging domain without the need for computationally expensive full fine-tuning. \review{It should be noted that the efficiency discussed above refers specifically to parameter efficiency during adaptation, not inference speed.} 

\subsection{Limitations and Future Work}
Despite its strengths, CFR-SAM has limitations that provide directions for future research. First, its performance on rare cell types (e.g., Dead cells, F1-score: 45\%) remains lower than on common types (e.g., Neoplastic cells, F1-score: 76\%), likely due to imbalanced annotation in datasets. Future work could explore class-balanced sampling or synthetic data augmentation to improve rare-class generalization. Second, the performance upper bound is partly constrained by the accuracy of the first-stage prompt learner. In extremely dense nuclear regions, minor inaccuracies in prompt points can lead to instance merging or missed splits. Future work could explore more discriminative prompt generation networks or iterative refinement mechanisms.

\section{Conclusion}
In this paper, we present CFR-SAM, a parameter-efficient fine-tuning framework that successfully adapts the Segment Anything Model for the challenging task of nucleus instance segmentation. We identify and address three core limitations of SAM in this domain: its lack of local feature sensitivity, underutilization of shallow encoder features, and the boundary ambiguity in its decoder outputs. The proposed MALA module effectively bridges the gap between SAM's global prior and the need for local precision by injecting adaptive, multi-scale contextual information. The HMFM ensures that low-level details are fully leveraged, while the BGMR explicitly enhances boundary awareness through a dedicated refinement process. Collectively, these components enable CFR-SAM to generate high-quality instance masks directly, without relying on complex post-processing pipelines.

Comprehensive evaluations on multiple public datasets confirm that our method sets a new state-of-the-art, achieving superior segmentation accuracy and generalization capability. Notably, this performance gain is attained with a minimal increase in trainable parameters, highlighting the efficiency of our design. This study highlights the significant potential of parameter-efficient adaptation strategies in harnessing powerful foundation models for specialized medical imaging tasks, offering a practical path forward for clinical application. Future work will focus on improving the robustness of prompt generation in extremely dense regions and enhancing the model's performance on rare cell types through advanced data augmentation strategies.

\bibliographystyle{elsarticle-harv} 
\bibliography{reference}

\end{document}

%% file: fft_motivation.tex
\begin{figure}[t]
    \centering
    \footnotesize
    \begin{tabular}{@{}c@{\hspace{0.5mm}}c@{\hspace{0.5mm}}c@{\hspace{0.5mm}}c@{\hspace{0.5mm}}c@{}}
        \rotatebox{90}{\makebox[1.9cm][c]{GT}} &
        \includegraphics[width=0.11\textwidth,height=1.9cm]{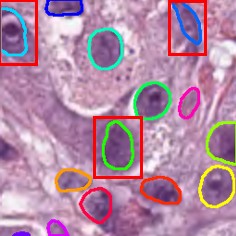} &
        \includegraphics[width=0.11\textwidth,height=1.9cm]{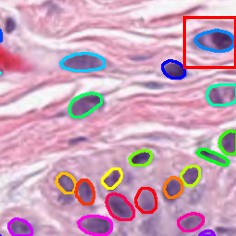} &
        \includegraphics[width=0.11\textwidth,height=1.9cm]{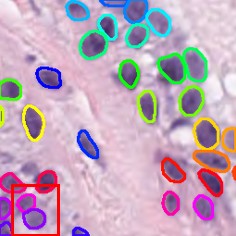} &
        \includegraphics[width=0.11\textwidth,height=1.9cm]{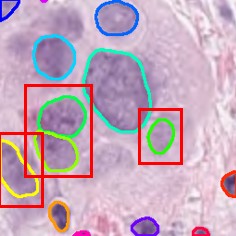} \\
        
        \rotatebox{90}{\makebox[1.9cm][c]{SAM}} &
        \includegraphics[width=0.11\textwidth,height=1.9cm]{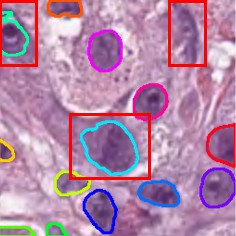} &
        \includegraphics[width=0.11\textwidth,height=1.9cm]{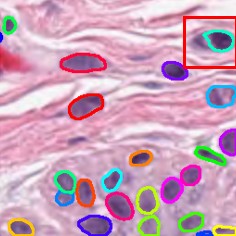} &
        \includegraphics[width=0.11\textwidth,height=1.9cm]{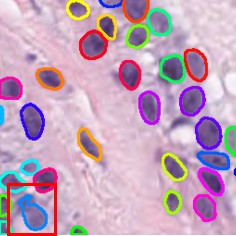} &
        \includegraphics[width=0.11\textwidth,height=1.9cm]{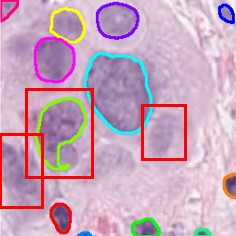} \\
        
        \rotatebox{90}{\makebox[1.9cm][c]{Ours}} &
        \includegraphics[width=0.11\textwidth,height=1.9cm]{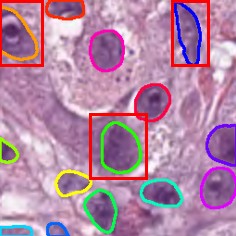} &
        \includegraphics[width=0.11\textwidth,height=1.9cm]{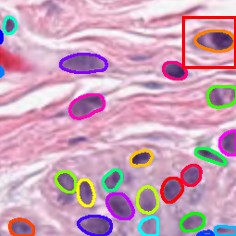} &
        \includegraphics[width=0.11\textwidth,height=1.9cm]{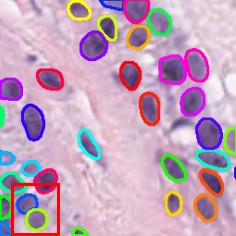} &
        \includegraphics[width=0.11\textwidth,height=1.9cm]{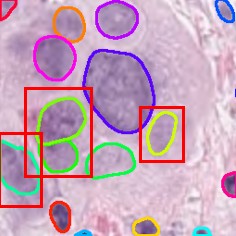} \\
    \end{tabular}
    
    \caption{This figure illustrates our motivation. Compared to the segmentation results of a full fine-tuned SAM (second row), our CFR-SAM achieves superior precision, particularly in capturing faint boundaries and distinguishing between densely clustered objects.}
    \label{fig:motivation}
\end{figure}

%% file: Pipeline.tex
\begin{figure*}[t]
    \centering
    \includegraphics[width=1\textwidth]{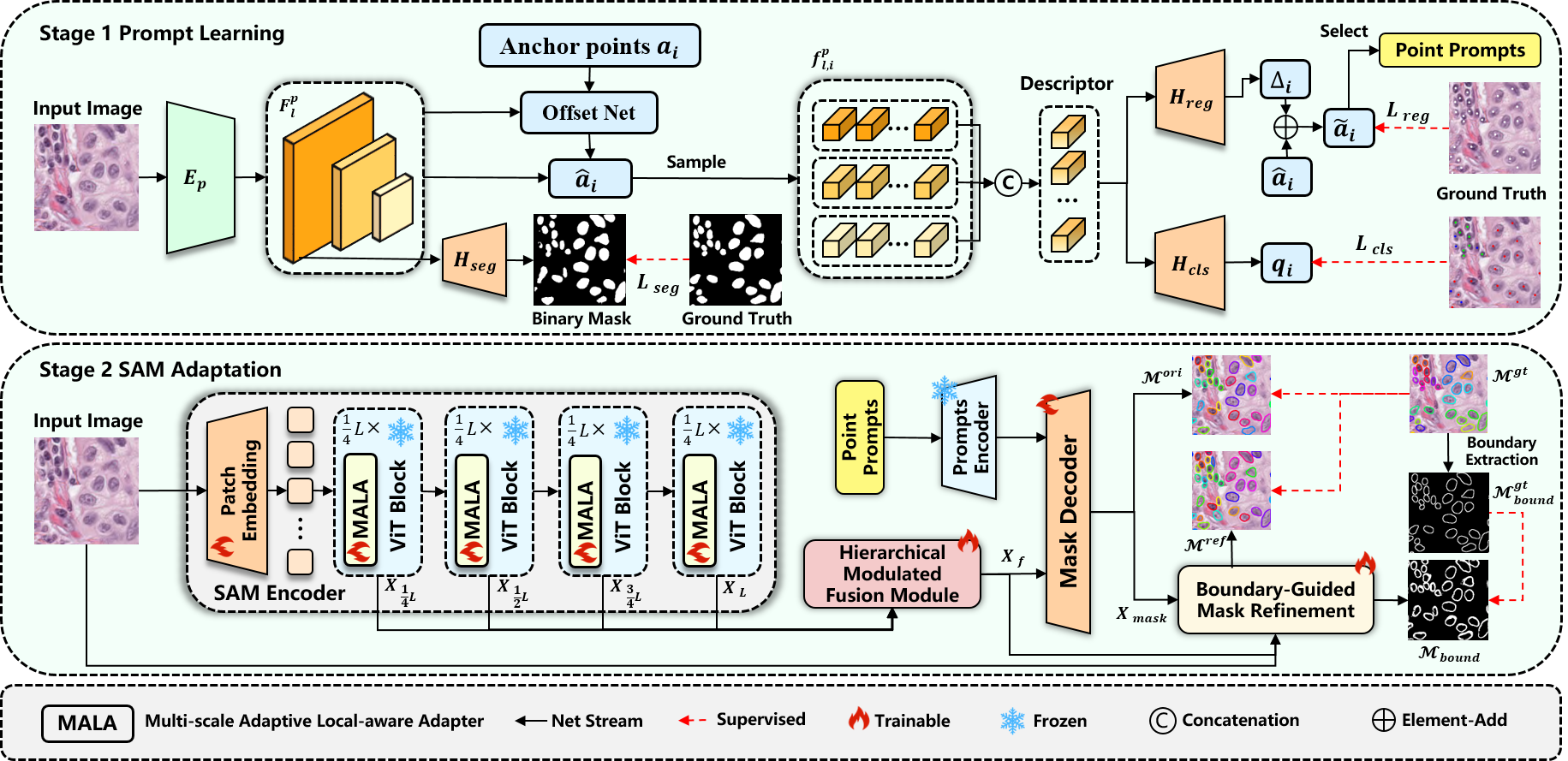}
    \caption{The pipeline illustrates the two-stage procedure of CFR-SAM. First, a Prompt Learner predicts nuclei centroids to generate point prompts. Second, a PEFT strategy adapts the SAM for nuclear instance segmentation. The final instance segmentation results are displayed with contours overlaid on each detected nucleus for clarity.}

    \label{fig:pipeline}
\end{figure*}

%% file: Module1.tex
\begin{figure}[!t]
    \centering
    \includegraphics[width=0.5\textwidth]{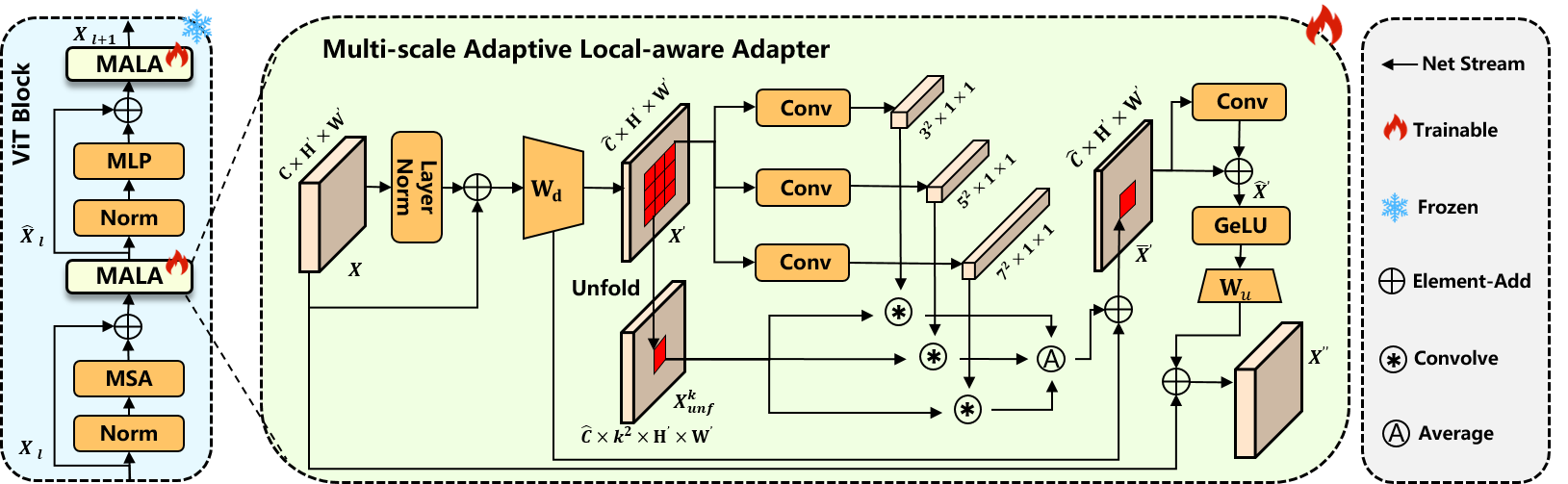}
    \caption{The structure of the proposed MALA is detailed, showcasing its role in enhancing the frozen Transformer backbone.}

    \label{fig:module1}
\end{figure}

%% file: Module2.tex
\begin{figure}[t]
    \centering
    \includegraphics[width=0.48\textwidth]{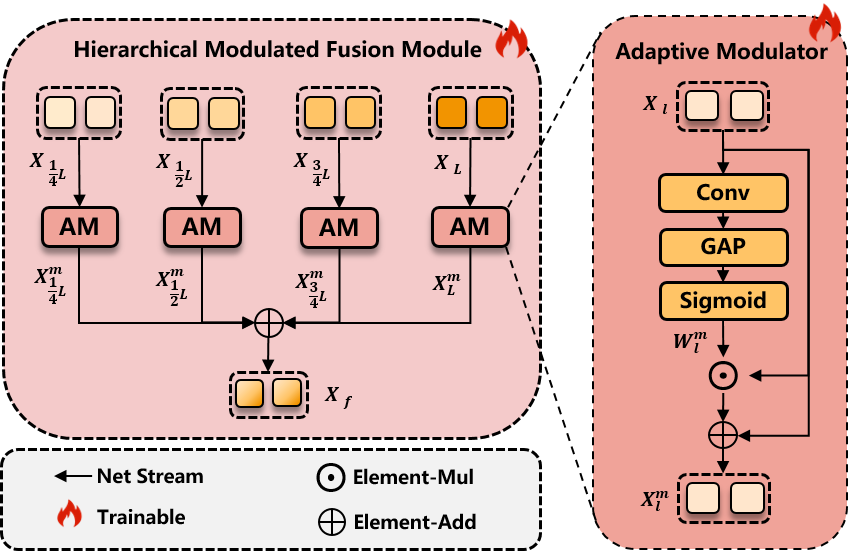}
    \caption{The design of the HMFM is presented, highlighting its mechanism for aggregating multi-level encoder features.}

    \label{fig:module2}
\end{figure}

%% file: Module3.tex
\begin{figure*}[!ht]
    \centering
    \includegraphics[width=1\textwidth]{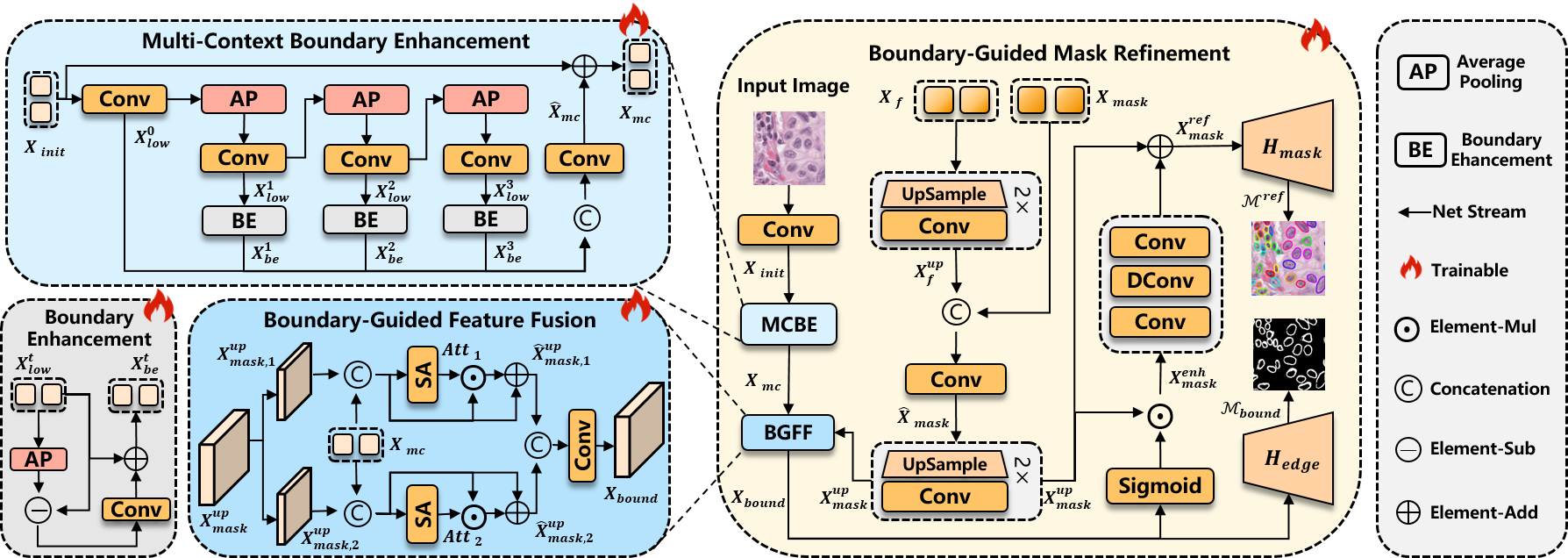}
    \caption{The workflow of the BGMR is depicted, demonstrating the process of integrating boundary cues to refine the initial mask prediction.}

    \label{fig:module3}
\end{figure*}

%% file: pannuke_instance.tex
\begin{table*}[!ht]
\renewcommand{\arraystretch}{1.1}
\centering
\caption{Performance comparison on the PanNuke dataset for nuclei instance segmentation. Following \citep{chen2023cpp, horst2024cellvit}, both binary PQ (bPQ) and multi-class PQ (mPQ) are computed for evaluation. \textcolor{red}{Red}/\textcolor{blue}{Blue} represents the best/second performance. Params* denotes trainable parameters of model.}
\resizebox{2\columnwidth}{!}{%
\begin{tabular}{l|cc|cc|cc|cc|cc|cc|cc}
\toprule[2pt]

\multicolumn{1}{l|}{\multirow{2}{*}{Model}} & \multicolumn{2}{c|}{Mask R-CNN}                 & \multicolumn{2}{c|}{HoVer-Net}                & \multicolumn{2}{c|}{CPP-Net}       & \multicolumn{2}{c|}{PointNu-Net}           & \multicolumn{2}{c|}{CellViT-H}                & \multicolumn{2}{c|}{CFR-SAM-B} & \multicolumn{2}{c}{CFR-SAM-H} \\
\multicolumn{1}{l|}{}                       & \multicolumn{2}{c|}{\citep{he2017mask}}  & \multicolumn{2}{c|}{\citep{graham2019hover}} & \multicolumn{2}{c|}{\citep{chen2023cpp}} & \multicolumn{2}{c|}{\citep{yao2023pointnu}} & \multicolumn{2}{c|}{\citep{horst2024cellvit}} & \multicolumn{2}{c|}{Ours}      & \multicolumn{2}{c}{Ours}      \\ \hline

Params* (M)  & \multicolumn{2}{c|}{41.9}       & \multicolumn{2}{c|}{122.8}    & \multicolumn{2}{c|}{37.6}      & \multicolumn{2}{c|}{122.8}   & \multicolumn{2}{c|}{699.7}       & \multicolumn{2}{c|}{63.4}          & \multicolumn{2}{c}{74.2}   \\ \hline
Tissue               & bPQ            & mPQ            & bPQ           & mPQ           & bPQ            & mPQ           & bPQ           & mPQ          & bPQ             & mPQ            & bPQ            & mPQ           & bPQ          & mPQ         \\ \hline
Adrenal              & 0.5546         & 0.3470         & 0.6962         & 0.4812        & 0.7031        & 0.4922        & 0.7134        & 0.5115       & 0.7086         & \textcolor{blue}{0.5134}        & \textcolor{blue}{0.7161}           & 0.5111        & \textcolor{red}{0.7195}       & \textcolor{red}{0.5140}      \\

Bile Duct            & 0.5567         & 0.3536         & 0.6696         & 0.4714        & 0.6739        & 0.4650        & 0.6814        & 0.4868       & 0.6784         & 0.4887        & \textcolor{blue}{0.6931}           & \textcolor{blue}{0.4952}        & \textcolor{red}{0.6935}       & \textcolor{red}{0.4953}      \\

Bladder              & 0.6049         & 0.5065         & 0.7031         & 0.5792        & 0.7057        & \textcolor{blue}{0.5932}      & 0.7226        & \textcolor{red}{0.6065}          & 0.7068         & 0.5844        & \textcolor{blue}{0.7234}           & 0.5902        & \textcolor{red}{0.7252}       & 0.5920      \\

Breast               & 0.5574         & 0.3882         & 0.6470         & 0.4902        & 0.6718        & 0.5066       & 0.6709        & 0.5147       & 0.6748         & 0.5180        & \textcolor{blue}{0.6781}           & \textcolor{blue}{0.5261}        & \textcolor{red}{0.6804}       & \textcolor{red}{0.5280}      \\

Cervix               & 0.5483         & 0.3402         & 0.6652         & 0.4438        & 0.6880        & 0.4779       & 0.6899        & 0.5014       & 0.6872         & 0.4984        & \textcolor{blue}{0.7032}           & \textcolor{blue}{0.5092}        & \textcolor{red}{0.7073}       & \textcolor{red}{0.5123}      \\

Conlon               & 0.4603         & 0.3122         & 0.5575         & 0.4095        & 0.5888        & 0.4269       & 0.5945        & 0.4509       & 0.5921         & 0.4485        & \textcolor{blue}{0.6080}           & \textcolor{blue}{0.4681}        & \textcolor{red}{0.6113}       & \textcolor{red}{0.4706}      \\

Esophagus            & 0.5691         & 0.4311         & 0.6427         & 0.5085        & 0.6755        & 0.5410       & 0.6766        & 0.5504       & 0.6682         & 0.5454        & \textcolor{blue}{0.6895}           & \textcolor{blue}{0.5635}        & \textcolor{red}{0.6919}       & \textcolor{red}{0.5660}      \\

Head Neck            & 0.5457         & 0.3946         & 0.6331         & 0.4530        & 0.6468        & 0.4667       & 0.6546        & 0.4838       & 0.6544         & 0.4913        & \textcolor{blue}{0.6694}           & \textcolor{blue}{0.5093}        & \textcolor{red}{0.6761}       & \textcolor{red}{0.5141}      \\

Kidney               & 0.5092         & 0.3553         & 0.6836        & 0.4424         & 0.7001       & 0.5092        & 0.6912        & 0.5066       & 0.7092         & 0.5366        & \textcolor{blue}{0.7107}           & \textcolor{blue}{0.5673}        & \textcolor{red}{0.7136}       & \textcolor{red}{0.5688}      \\

Liver                & 0.6085         & 0.4103         & 0.7248         & 0.4974        & 0.7271        & 0.5099       & 0.7314        & 0.5174       & 0.7322         & 0.5224        & \textcolor{blue}{0.7415}           & \textcolor{blue}{0.5267}        & \textcolor{red}{0.7439}       & \textcolor{red}{0.5272}      \\

Lung                 & 0.5134         & 0.3182         & 0.6302         & 0.4004        & 0.6364        & 0.4234       & 0.6352        & 0.4048       & 0.6426         & 0.4314        & \textcolor{blue}{0.6549}           & \textcolor{blue}{0.4372}        & \textcolor{red}{0.6598}       & \textcolor{red}{0.4393}      \\

Ovarian              & 0.5784         & 0.4337         & 0.6309         & 0.4863        & 0.6792        & 0.5276       & \textcolor{blue}{0.6863}        & \textcolor{blue}{0.5484}       & 0.6722         & 0.5390        & 0.6853           & 0.5465        & \textcolor{red}{0.6871}       & \textcolor{red}{0.5488}      \\

Pancreatic           & 0.5460         & 0.3624         & 0.6491         & 0.4600        & 0.6742        & 0.4680       & 0.6791        & 0.4804       & 0.6658         & 0.4719        & \textcolor{blue}{0.6903}           & \textcolor{blue}{0.5073}        & \textcolor{red}{0.6925}       & \textcolor{red}{0.5095}      \\

Prostate             & 0.5789         & 0.3959         & 0.6615         & 0.5101        & 0.6903        & 0.5261       & 0.6854        & 0.5127       & 0.6821         & 0.5321        & \textcolor{blue}{0.6963}           & \textcolor{blue}{0.5422}        & \textcolor{red}{0.6985}       & \textcolor{red}{0.5439}      \\

Skin                 & 0.5021         & 0.2665         & 0.6234         & 0.3429        & 0.6192        & 0.3547       & 0.6494        & 0.4011       & 0.6565         & 0.4339        & \textcolor{blue}{0.6755}           & \textcolor{blue}{0.4387}        & \textcolor{red}{0.6777}       & \textcolor{red}{0.4399}      \\

Stomack              & 0.5976         & 0.3684         & 0.6886         & \textcolor{red}{0.4726}        & 0.7043        & 0.4553      & 0.7010        & 0.4517        & 0.7022         & \textcolor{blue}{0.4705}        & \textcolor{blue}{0.7100}           & 0.4494        & \textcolor{red}{0.7155}       & 0.4543      \\

Testis               & 0.5420         & 0.3512         & 0.6890         & 0.4754        & 0.7006        & 0.4917       & 0.7058        & \textcolor{blue}{0.5334}        & 0.6955         & 0.5127        & \textcolor{blue}{0.7108}           & 0.5312        & \textcolor{red}{0.7157}       & \textcolor{red}{0.5343}      \\

Thyroid              & 0.5712         & 0.3037         & 0.6983         & 0.4315        & 0.7094        & 0.4344       & 0.7076        & 0.4508        & 0.7151         & 0.4519        & \textcolor{blue}{0.7274}           & \textcolor{blue}{0.4652}        & \textcolor{red}{0.7293}       & \textcolor{red}{0.4662}      \\

Uterus               & 0.5589         & 0.3683         & 0.6393         & 0.4393        & 0.6622        & 0.4790       & 0.6634        & 0.4846        & 0.6625         & 0.4737        & \textcolor{blue}{0.6730}           & \textcolor{blue}{0.4905}        & \textcolor{red}{0.6754}       & \textcolor{red}{0.4915}            \\ \hline

Average              & 0.5528         & 0.3688         & 0.6596         & 0.4629        & 0.6767        & 0.4817       & 0.6808        & 0.4957        & 0.6793         & 0.4980        & \textcolor{blue}{0.6924}           & \textcolor{blue}{0.5092}        & \textcolor{red}{0.6955}       & \textcolor{red}{0.5114}      \\

Std                  & 0.0370         & 0.0543         & 0.0036         & 0.0076        & -        & -       & -        & -       & 0.0318         & 0.0413        & 0.0301           & 0.0437        & 0.0298       & 0.0436      \\ \bottomrule[2pt]
\end{tabular}}

\label{tab:pannuke_instance}
\end{table*}

%% file: statistic.tex
\definecolor{mygray}{gray}{.9}
\begin{table}[t]
\caption{Performance of our CFR-SAM on PanNuke over 5 independent runs.}
\renewcommand{\arraystretch}{1.1}
\centering
\small

\resizebox{1\columnwidth}{!}{
\review{
\begin{tabular}{lcccc}
\toprule[2pt]
\multirow{2}{*}{Model}        & Trainable            & \multirow{2}{*}{mPQ} & \multirow{2}{*}{bPQ} & \multirow{2}{*}{AJI}\\
                               & Params (M)       &                      &           &           \\ \midrule  
CellViT-H    & 699.7     & 0.4980                 & 0.6793               & 0.6519        \\
Ours-B       &  63.4     & 0.5091 $\pm$ 0.0013    & 0.6923 $\pm$ 0.0020  & 0.6865 $\pm$ 0.0001      \\
Ours-H       & 74.2      & 0.5115 $\pm$ 0.0003    & 0.6956 $\pm$ 0.0004  & 0.6887 $\pm$ 0.0004     \\ \bottomrule[2pt]
\end{tabular}}
}
\label{tab:statistic}
\end{table}

%% file: pannuke_det_cls.tex
\begin{table*}[!ht]
\centering
\renewcommand{\arraystretch}{1.1}
\caption{Comparison of Precision (P), Recall (R), and $F_1$-score ($F_1$) for nuclei detection and classification across all categories. Results are averaged over the three PanNuke splits.}
\resizebox{2\columnwidth}{!}{%
\begin{tabular}{lcccccccccccccccccc}
\toprule[2pt]
\centering
\multirow{3}{*}{Model} & \multicolumn{3}{c}{\multirow{2}{*}{Detection}} & \multicolumn{15}{c}{Classification}                                                                                                                            \\ \cline{5-19} 
                       & \multicolumn{3}{c}{}                           & \multicolumn{3}{c}{Neoplastic} & \multicolumn{3}{c}{Epithelial} & \multicolumn{3}{c}{Inflammatory} & \multicolumn{3}{c}{Connective} & \multicolumn{3}{c}{Dead} \\ \cline{2-19} 
                       & P              & R             & $F_1$            & P        & R        & $F_1$       & P        & R        & $F_1$       & P         & R         & $F_1$       & P        & R        & $F_1$       & P      & R      & $F_1$     \\ \hline
DIST \citep{naylor2018segmentation}                  & 0.74           & 0.71          & 0.73          & 0.49     & 0.55     & 0.50     & 0.38     & 0.33     & 0.35     & 0.42      & 0.45      & 0.42     & 0.42     & 0.37     & 0.39     & 0.00   & 0.00   & 0.00   \\
Mask-RCNN  \citep{he2017mask}            & 0.76           & 0.68          & 0.72          & 0.55     & 0.63     & 0.59     & 0.52     & 0.52     & 0.52     & 0.46      & 0.54      & 0.50     & 0.42     & 0.43     & 0.42     & 0.17   & 0.30   & 0.22   \\
Micro-Net \citep{raza2019micro}             & 0.78           & \textcolor{blue}{0.82}          & 0.80          & 0.59     & 0.66     & 0.62     & 0.63     & 0.54     & 0.58     & 0.59      & 0.46      & 0.52     & 0.50     & 0.45     & 0.47     & 0.23   & 0.17   & 0.19   \\
HoVer-Net \citep{graham2019hover}             & 0.82           & 0.79          & 0.80          & 0.58     & 0.67     & 0.62     & 0.54     & 0.60     & 0.56     & 0.56      & 0.51      & 0.54     & 0.52     & 0.47     & 0.49     & 0.28   & 0.35   & 0.31   \\
STARDIST \citep{schmidt2018cell}              & \textcolor{blue}{0.85}           & 0.80          & \textcolor{blue}{0.82}          & 0.69     & 0.69     & 0.69     & 0.73     & 0.68     & 0.70     & 0.62      & 0.53      & 0.57     & 0.54     & 0.49     & 0.51     & 0.39   & 0.09   & 0.10   \\
CPP-Net \citep{chen2023cpp}               & \textcolor{red}{0.87}           & 0.78          & \textcolor{blue}{0.82}          & \textcolor{blue}{0.74}     & 0.67     & 0.70     & 0.74     & 0.70     & 0.72     & 0.60      & \textcolor{blue}{0.57}      & 0.58     & 0.57     & 0.49     & 0.53     & 0.41   & \textcolor{blue}{0.36}   & \textcolor{blue}{0.38}   \\

PointNu-Net \citep{yao2023pointnu}        & 0.81  &0.81   &0.81   & \textcolor{blue}{0.74}   & \textcolor{blue}{0.72}   & \textcolor{blue}{0.73}   & \textcolor{blue}{0.75}   &0.71   & \textcolor{blue}{0.73}   & \textcolor{blue}{0.63}   &\textcolor{blue}{0.57}   & \textcolor{blue}{0.60}   & \textcolor{blue}{0.61}   &\textcolor{blue}{0.55}   & \textcolor{blue}{0.58}   & \textcolor{blue}{0.45}   &0.24   &0.31 \\

CellViT-H \citep{horst2024cellvit}             & 0.84           & 0.81          & \textcolor{red}{0.83}          & 0.72     & 0.69     & 0.71     & 0.72     & \textcolor{blue}{0.73}     & \textcolor{blue}{0.73}     & 0.59      & \textcolor{blue}{0.57}      & 0.58     & 0.55     & 0.52     & 0.53     & 0.43   & 0.32   & 0.36   \\

Ours-H                   & 0.82               & \textcolor{red}{0.84}              & \textcolor{red}{0.83}              & \textcolor{red}{0.75}         & \textcolor{red}{0.78}         & \textcolor{red}{0.76}         & \textcolor{red}{0.77}         & \textcolor{red}{0.80}         & \textcolor{red}{0.78}         & \textcolor{red}{0.69}          & \textcolor{red}{0.71}          & \textcolor{red}{0.70}         & \textcolor{red}{0.63}         & \textcolor{red}{0.62}         & \textcolor{red}{0.62}         & \textcolor{red}{0.49}       & \textcolor{red}{0.42}       & \textcolor{red}{0.45}       \\ \bottomrule[2pt]
\end{tabular}}

\label{tab:pannuke_detect}
\end{table*}

%% file: pannuke_mpq.tex
\begin{table}[t]
\centering              
\renewcommand{\arraystretch}{1.1}
\caption{Average PQ for each nuclear category across the three PanNuke splits. N, E, I, C, and D represent Neoplastic, Epithelial, Inflammatory, Connective, and Dead, respectively.}
\begin{tabular}{lccccc}
\toprule[2pt]
Model & N & E & I & C & D \\ \midrule
DIST        & 0.439 & 0.290 & 0.343 & 0.275 & 0.000 \\
Mask-RCNN   & 0.472 & 0.403 & 0.290 & 0.300 & 0.069 \\
Micro-Net   & 0.504 & 0.442 & 0.333 & 0.334 & 0.051 \\
HoVer-Net  & 0.551 & 0.491 & 0.417 & 0.388 & 0.139 \\
STARDIST   & 0.547 & 0.532 & 0.424 & 0.380 & 0.123 \\
CPP-Net  & 0.571 & 0.565 & 0.405 & 0.395 & 0.131 \\
PointNu-Net & 0.579  & 0.577    & \textcolor{red}{0.433}  & 0.409    & \textcolor{red}{0.154} \\
CellViT-H   & \textcolor{blue}{0.581} & \textcolor{blue}{0.583} & 0.417 & \textcolor{blue}{0.423} & 0.149 \\
Ours-H        & \textcolor{red}{0.599} & \textcolor{red}{0.589} & \textcolor{blue}{0.428} & \textcolor{red}{0.433} & \textcolor{blue}{0.153} \\ \bottomrule[2pt]
\end{tabular}
\label{tab:pannuke_mpq}
\end{table}

%% file: h_compare.tex
\begin{figure}[!t]
    \centering
    \footnotesize
    \begin{tabular}{@{}c@{\hspace{0.5mm}}c@{\hspace{0.5mm}}c@{\hspace{0.5mm}}c@{\hspace{0.5mm}}c@{}}
        \rotatebox{90}{\makebox[1.9cm][c]{GT}} &
        \includegraphics[width=0.11\textwidth,height=1.9cm]{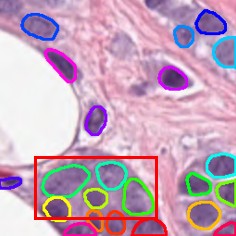} &
        \includegraphics[width=0.11\textwidth,height=1.9cm]{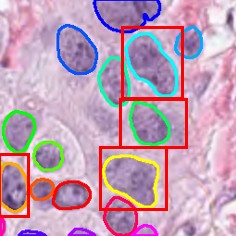} &
        \includegraphics[width=0.11\textwidth,height=1.9cm]{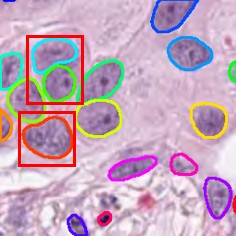} &
        \includegraphics[width=0.11\textwidth,height=1.9cm]{395_1.jpg} \\
        
        \rotatebox{90}{\makebox[1.9cm][c]{CellViT}} &
        \includegraphics[width=0.11\textwidth,height=1.9cm]{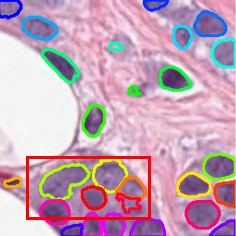} &
        \includegraphics[width=0.11\textwidth,height=1.9cm]{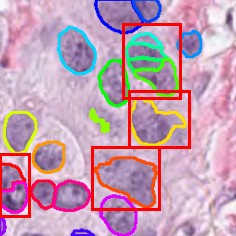} &
        \includegraphics[width=0.11\textwidth,height=1.9cm]{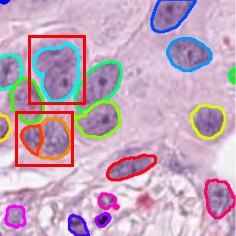} &
        \includegraphics[width=0.11\textwidth,height=1.9cm]{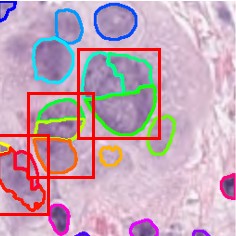} \\
        
        \rotatebox{90}{\makebox[1.9cm][c]{Ours}} &
        \includegraphics[width=0.11\textwidth,height=1.9cm]{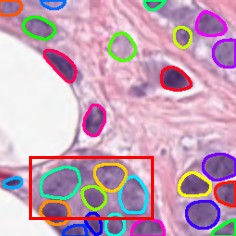} &
        \includegraphics[width=0.11\textwidth,height=1.9cm]{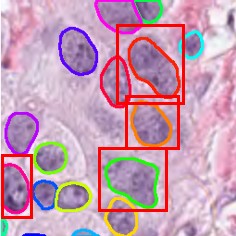} &
        \includegraphics[width=0.11\textwidth,height=1.9cm]{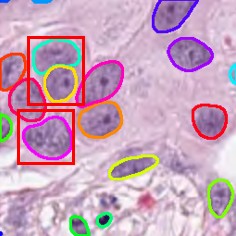} &
        \includegraphics[width=0.11\textwidth,height=1.9cm]{395_3.jpg} \\
    \end{tabular}
    
    \caption{Visual comparison of our method and CellViT on the PanNuke dataset for nuclei instance segmentation. Each predicted and GT instance mask is assigned a distinct color at random for visual separation.  The red boxes highlight the better results achieved by our model.}
    \label{fig:h_compare}
\end{figure}

%% file: instance_cls.tex
\begin{figure}[t]
    \centering
    \footnotesize
    \begin{tabular}{@{}c@{\hspace{0.5mm}}c@{\hspace{0.5mm}}c@{\hspace{0.5mm}}c@{\hspace{0.5mm}}c@{}}
        \rotatebox{90}{\makebox[1.9cm][c]{GT}} &
        \includegraphics[width=0.11\textwidth,height=1.9cm]{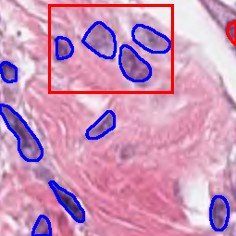} &
        \includegraphics[width=0.11\textwidth,height=1.9cm]{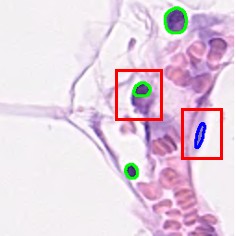} &
        \includegraphics[width=0.11\textwidth,height=1.9cm]{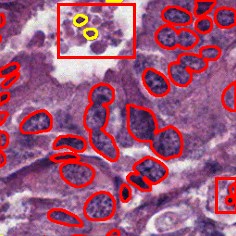} &
        \includegraphics[width=0.11\textwidth,height=1.9cm]{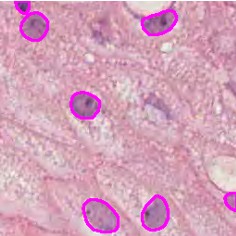} \\
        
        \rotatebox{90}{\makebox[1.9cm][c]{CellViT}} &
        \includegraphics[width=0.11\textwidth,height=1.9cm]{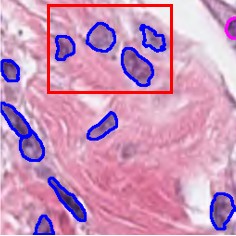} &
        \includegraphics[width=0.11\textwidth,height=1.9cm]{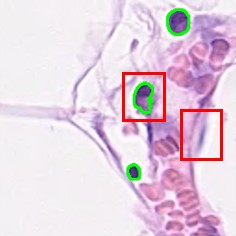} &
        \includegraphics[width=0.11\textwidth,height=1.9cm]{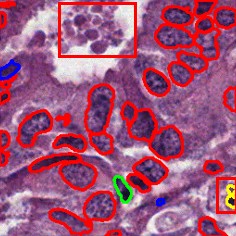} &
        \includegraphics[width=0.11\textwidth,height=1.9cm]{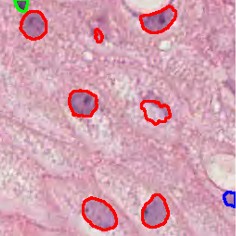} \\
        
        \rotatebox{90}{\makebox[1.9cm][c]{Ours}} &
        \includegraphics[width=0.11\textwidth,height=1.9cm]{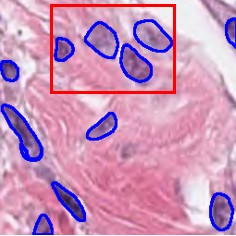} &
        \includegraphics[width=0.11\textwidth,height=1.9cm]{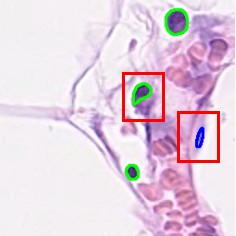} &
        \includegraphics[width=0.11\textwidth,height=1.9cm]{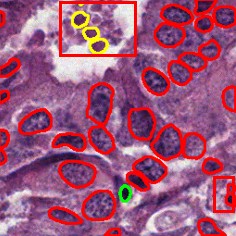} &
        \includegraphics[width=0.11\textwidth,height=1.9cm]{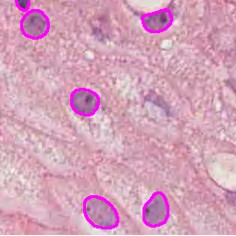} \\
    \end{tabular}

    \begin{center}
    \begin{tabular}{cccccccccc}
    \hspace{-0.3cm}
    \legendbox{color1} & \hspace{-0.3cm} \scalebox{0.9}{Neoplastic} & \hspace{-0.3cm}
    \legendbox{color5} & \hspace{-0.3cm} \scalebox{0.9}{Epithelial} & \hspace{-0.3cm}
    \legendbox{color2} & \hspace{-0.3cm} \scalebox{0.9}{Inflammatory} & \hspace{-0.3cm}
    \legendbox{color3} & \hspace{-0.3cm} \scalebox{0.9}{Connective} & \hspace{-0.3cm}
    \legendbox{color4} & \hspace{-0.3cm} \scalebox{0.9}{Dead}\\
\end{tabular}
\end{center}
    
    \caption{Visual comparison of our method and CellViT on the PanNuke dataset for nuclei classification. Nuclear types are represented by distinct boundary colors.}
    \label{fig:instance_cls_compare}
\end{figure}

%% file: cpm17.tex
\begin{table}[t]
\centering
\renewcommand{\arraystretch}{1.1}
\setlength{\tabcolsep}{12pt}
\caption{Performance comparison on the CPM-17 dataset for nuclei instance segmentation.}
\begin{tabular}{lcc}
\toprule[2pt]
Model & AJI          & PQ          \\ \hline
U-Net \citep{ronneberger2015u}                 & 0.666        & 0.625       \\
DCAN  \citep{chen2017dcan}                 & 0.561        & 0.545       \\
Mask-RCNN  \citep{he2017mask}            & 0.684        & 0.674       \\
DIST \citep{naylor2018segmentation}                  & 0.616        & 0.504       \\
Micro-Net  \citep{raza2019micro}            & 0.668        & 0.661       \\
Hover-Net \citep{graham2019hover}             & 0.705        & 0.697       \\
FEEDNet \citep{deshmukh2022feednet}               & 0.701        & 0.705       \\
HARU-Net \citep{chen2023enhancing}              & \textcolor{blue}{0.721}        & 0.701       \\
PointNu-Net \citep{yao2023pointnu}           & 0.712        & \textcolor{blue}{0.706}       \\
Ours-H                   & \textcolor{red}{0.725}        & \textcolor{red}{0.719}       \\ \bottomrule[2pt]
\end{tabular}

\label{tab:cpm17}
\end{table}

%% file: monuseg.tex
\begin{table}[t]
\centering
\renewcommand{\arraystretch}{1.1}
\setlength{\tabcolsep}{12pt}
\caption{Performance comparison on the MoNuSeg dataset for nuclei instance segmentation. $\dagger$ indicates our reproduced results.}
\begin{tabular}{lcc}
\toprule[2pt]
Model & AJI          & PQ           \\ \midrule
U-Net \citep{ronneberger2015u}                & 0.556        & 0.478        \\
Mask-RCNN \citep{he2017mask}             & 0.546        & 0.365        \\
DCAN \citep{chen2017dcan}                  & 0.477        & 0.488        \\
DIST \citep{naylor2018segmentation}                 & 0.559        & 0.443        \\
Micro-Net \citep{raza2019micro}             & 0.560        & 0.519        \\
CIA-Net \citep{zhou2019cia}               & 0.620        & 0.577        \\
Hover-Net \citep{graham2019hover}             & \textcolor{blue}{0.661}        & 0.590        \\
CDNet \citep{he2021cdnet}                 & 0.630        & \textcolor{blue}{0.634}        \\
TopeSeg \citep{he2023toposeg}               & 0.643        & 0.625        \\
CellViT-H$^{\dagger}$ \citep{horst2024cellvit}             & 0.644        & 0.490             \\
Ours-H                   & \textcolor{red}{0.668}        & \textcolor{red}{0.662}        \\ \bottomrule[2pt] 
\end{tabular}

\label{tab:monuseg}
\end{table}

%% file: ablation.tex
\begin{table}[]
\renewcommand{\arraystretch}{1.1}
\centering
\caption{Ablation study of different modules. All trainable parameters are from the second-stage fine-tuning of SAM. The symbol $*$ indicates that we increase the low-rank dimension to match the trainable parameters. }

\begin{tabular}{lccc}
\toprule[2pt]
\multirow{2}{*}{Model}        & Trainable            & \multirow{2}{*}{mPQ} & \multirow{2}{*}{bPQ} \\
                               & Params (M)       &                      &                      \\ \midrule 
\multicolumn{4}{l}{\textbf{Different\ Fine-Tuning\ Method} }                                  \\
FFT                            & 93.7                 & 0.4940               & 0.6754               \\
Head Only                      & 4.1                  & 0.4848               & 0.6645               \\
\rowcolor{gray!20} \textbf{Ours}                  & 8.3                  & \textbf{0.5046}      & \textbf{0.6912}      \\ \midrule
\multicolumn{4}{l}{\textbf{Multi-Scale\ Adaptive\ Local-Aware\ Adapter}}                                              \\
\review{w/o MALA}                       & \review{4.6}                  & \review{0.4980}               & \review{0.6824}       \\
MALA$_{static\ kernel}$        & 7.3                  & 0.5033               & 0.6895               \\
\review{MALA$^*_{static\ kernel}$}        & \review{8.8}                  & \review{0.5039}               & \review{0.6902}               \\
MALA$_{learnable\ kernel}$     & 8.3                  & 0.5046               & 0.6912               \\ \midrule
\multicolumn{4}{l}{\textbf{Hierarchical\ Modulated\ Fusion\ Module}}                          \\
w/o HMFM                       & 8.0                  & 0.5037               & 0.6899               \\
add                            & 8.0                  & 0.5039               & 0.6898               \\
concatenate                    & 8.3                  & 0.5040               & 0.6906               \\
average                        & 8.0                  & 0.5041               & 0.6902               \\ \midrule
\multicolumn{4}{l}{\textbf{Boundary-Guided\ Mask\ Refinement}}                                  \\
w/o BGMR                       & 8.0                  & 0.4990               & 0.6829               \\
\review{BGMR->MLP}             & \review{8.3}         & \review{0.5023}      & \review{0.6886}      \\
w/o MCBE                       & 8.3                  & 0.5037               & 0.6900               \\
w/o BGFF                       & 8.3                  & 0.5032               & 0.6890               \\
w/o Boundary Supervise         & 8.3                  & 0.5033               & 0.6891               \\ \bottomrule[2pt]
\end{tabular}

\label{tab:ablation}
\end{table}

%% file: fft_paper.tex
\begin{figure}[t]
    \centering
    \setlength{\tabcolsep}{0.5mm}
    \begin{tabular}{c m{0.18\linewidth} m{0.18\linewidth} m{0.18\linewidth} m{0.18\linewidth} m{0.18\linewidth}}
        
        \rotatebox{90}{GT} &
        \includegraphics[width=\linewidth]{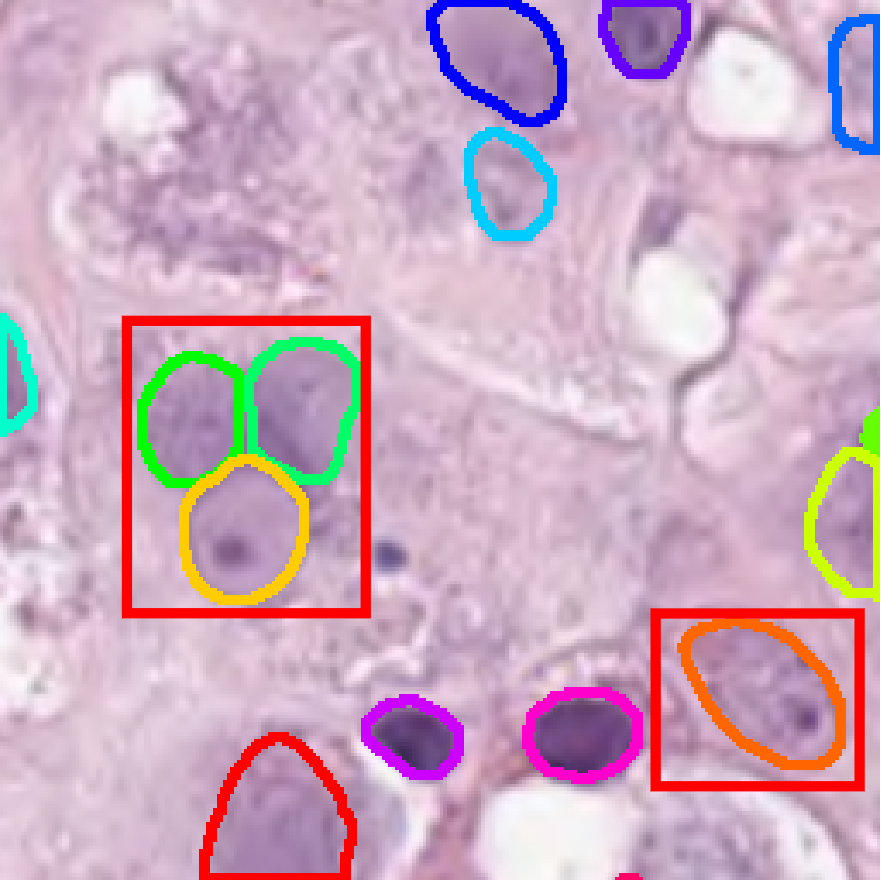} &
        \includegraphics[width=\linewidth]{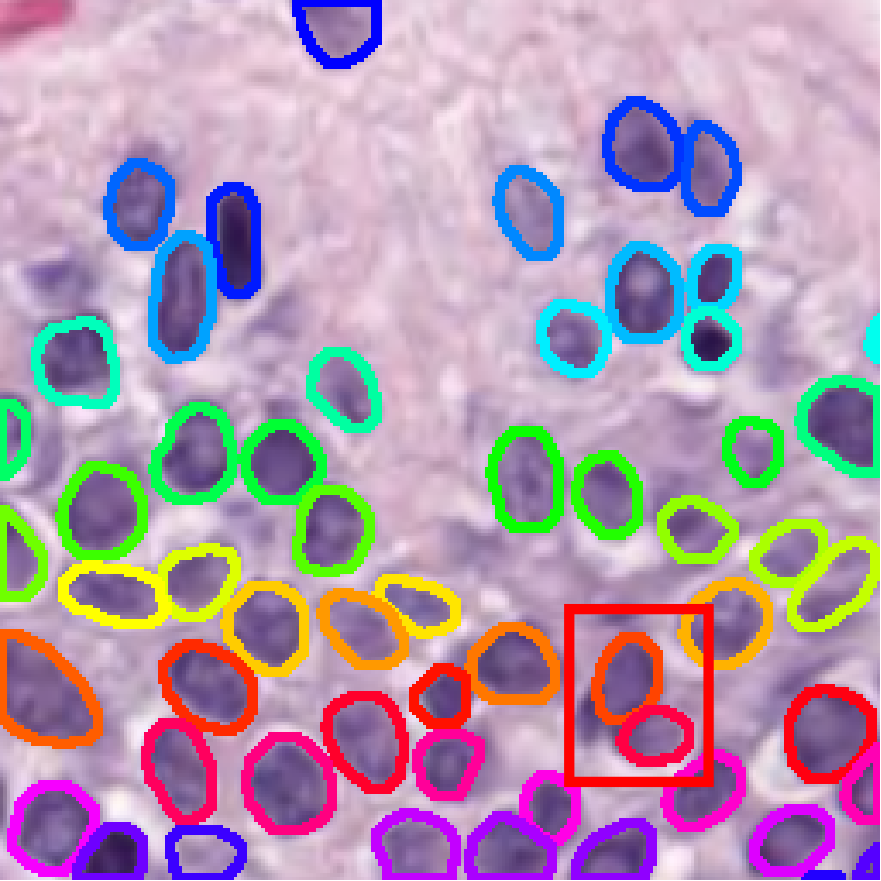} &
        \includegraphics[width=\linewidth]{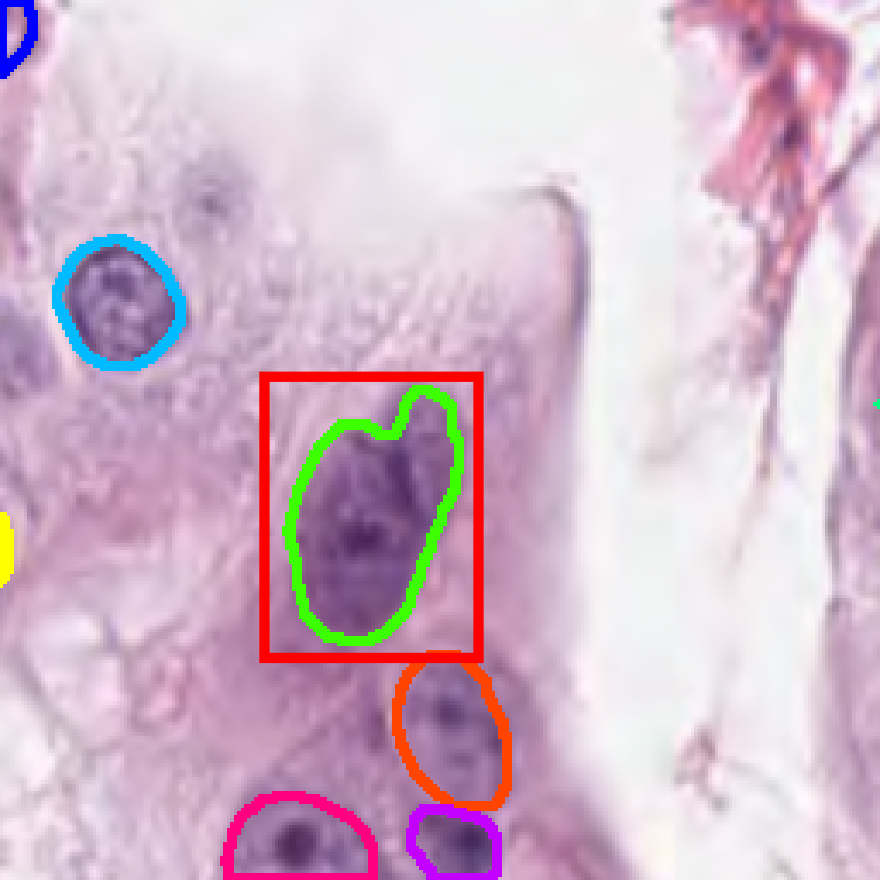} &
        \includegraphics[width=\linewidth]{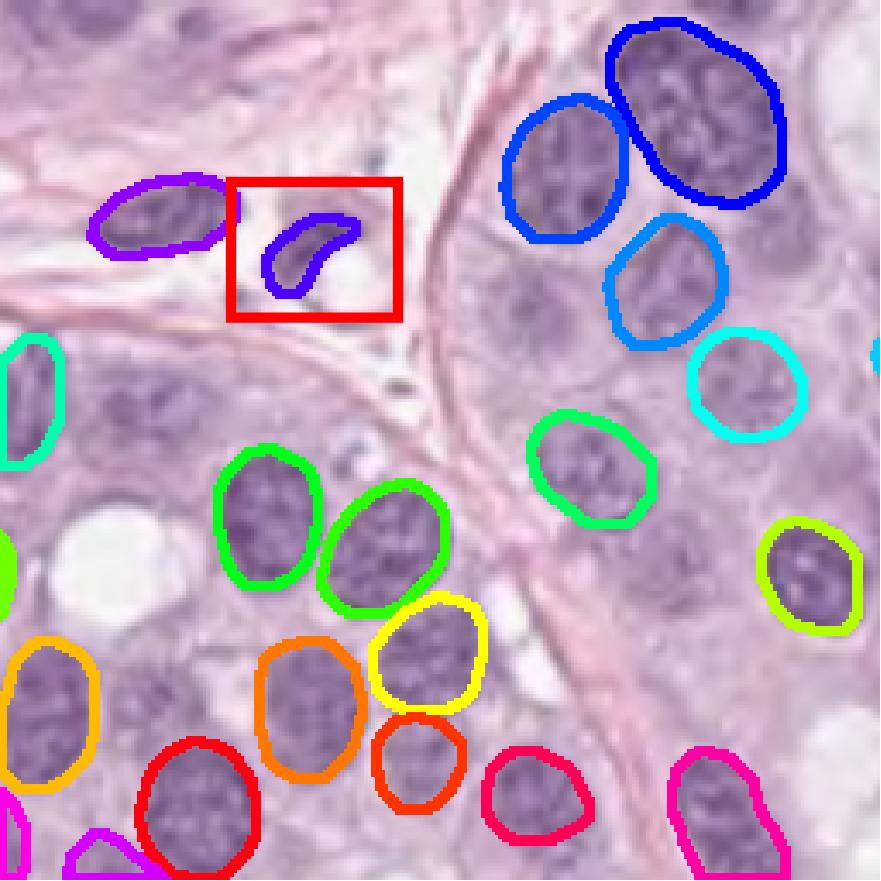} &
        \includegraphics[width=\linewidth]{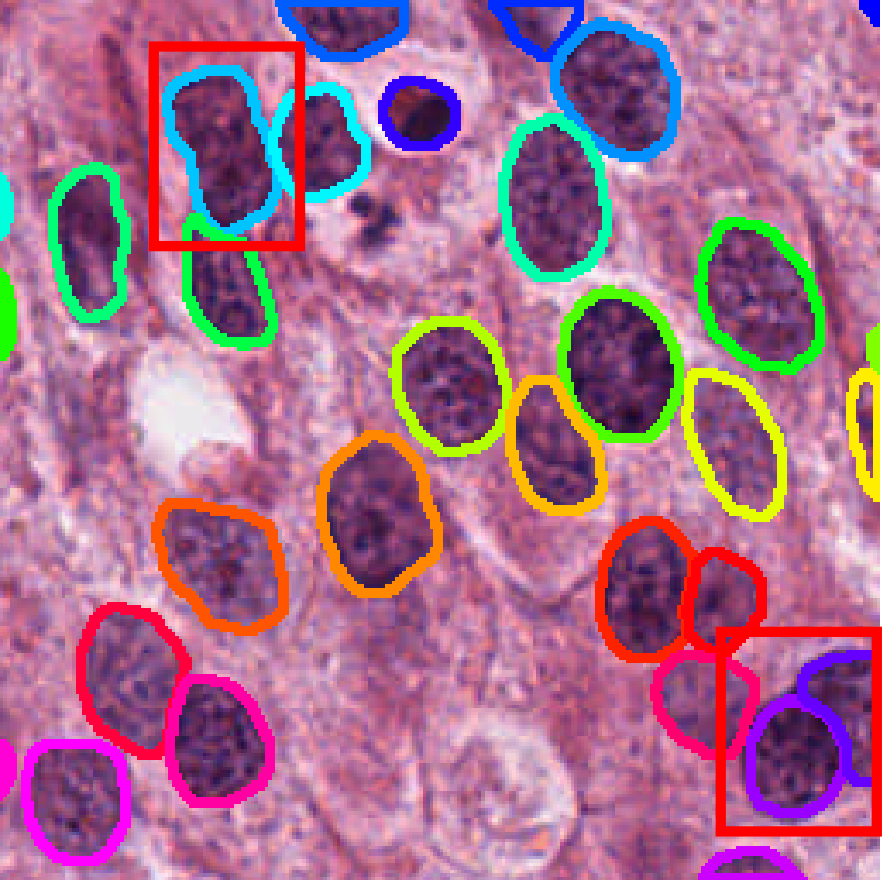} \\[1mm] %
        
        \rotatebox{90}{FFT} &
        \includegraphics[width=\linewidth]{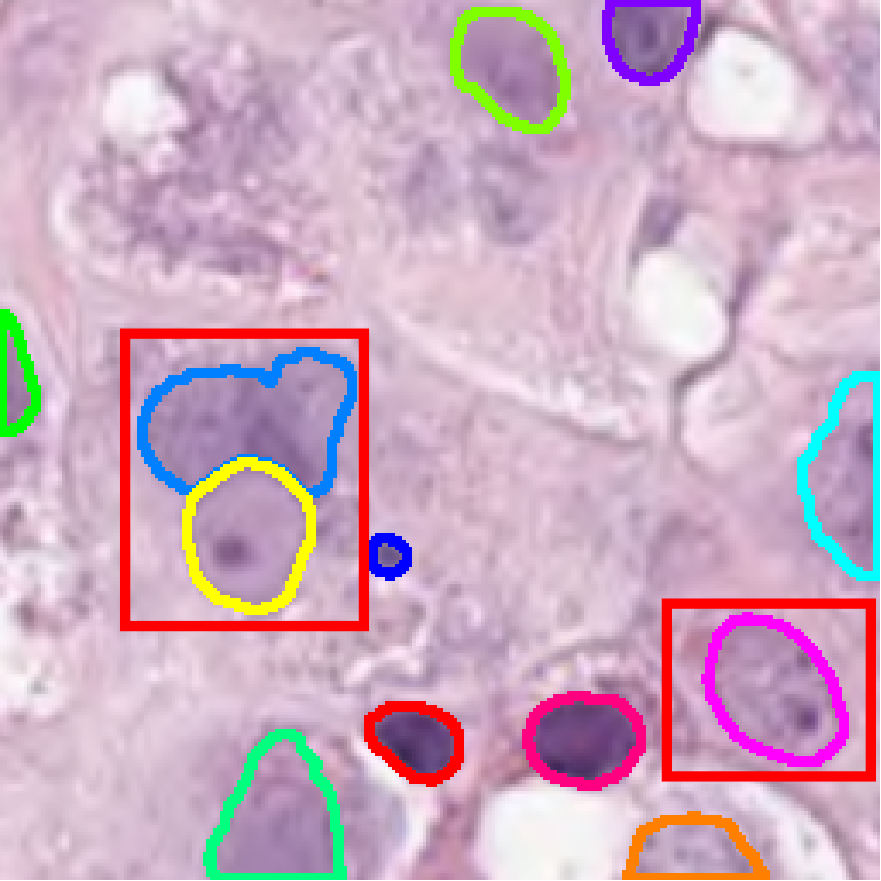} &
        \includegraphics[width=\linewidth]{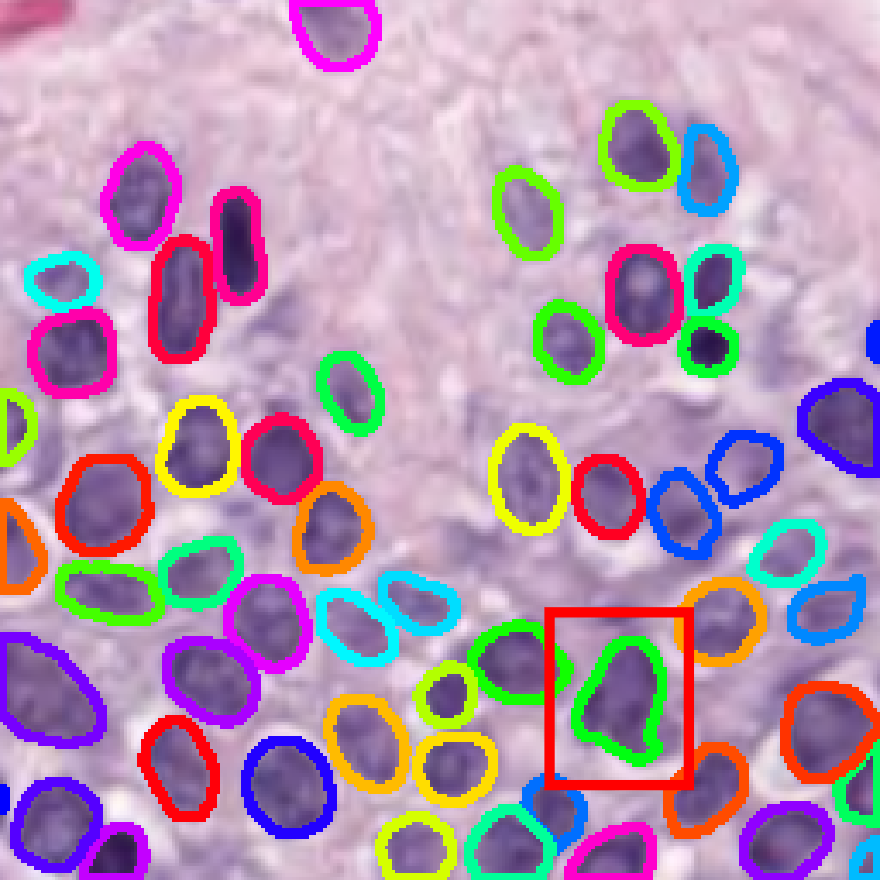} &
        \includegraphics[width=\linewidth]{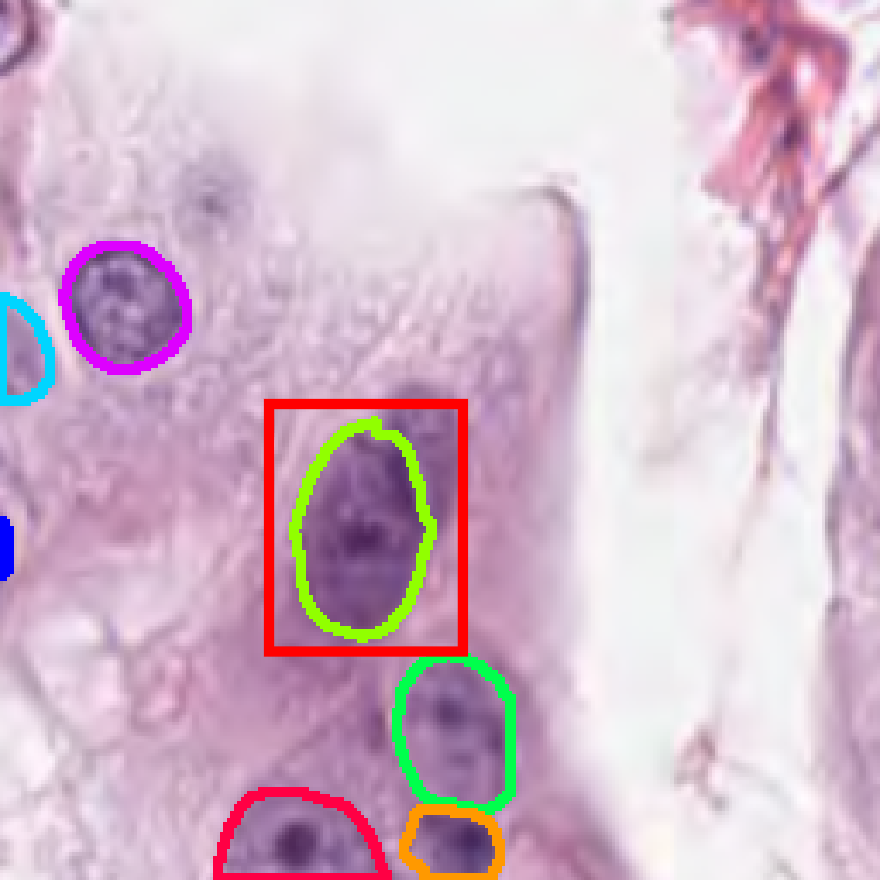} &
        \includegraphics[width=\linewidth]{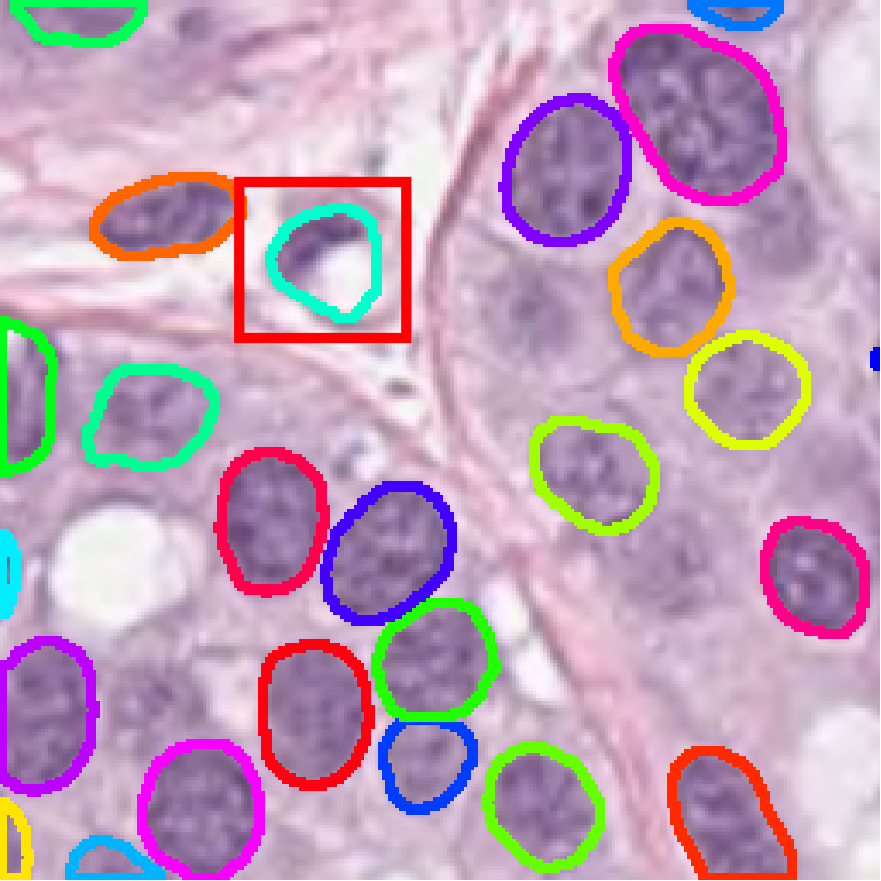} &
        \includegraphics[width=\linewidth]{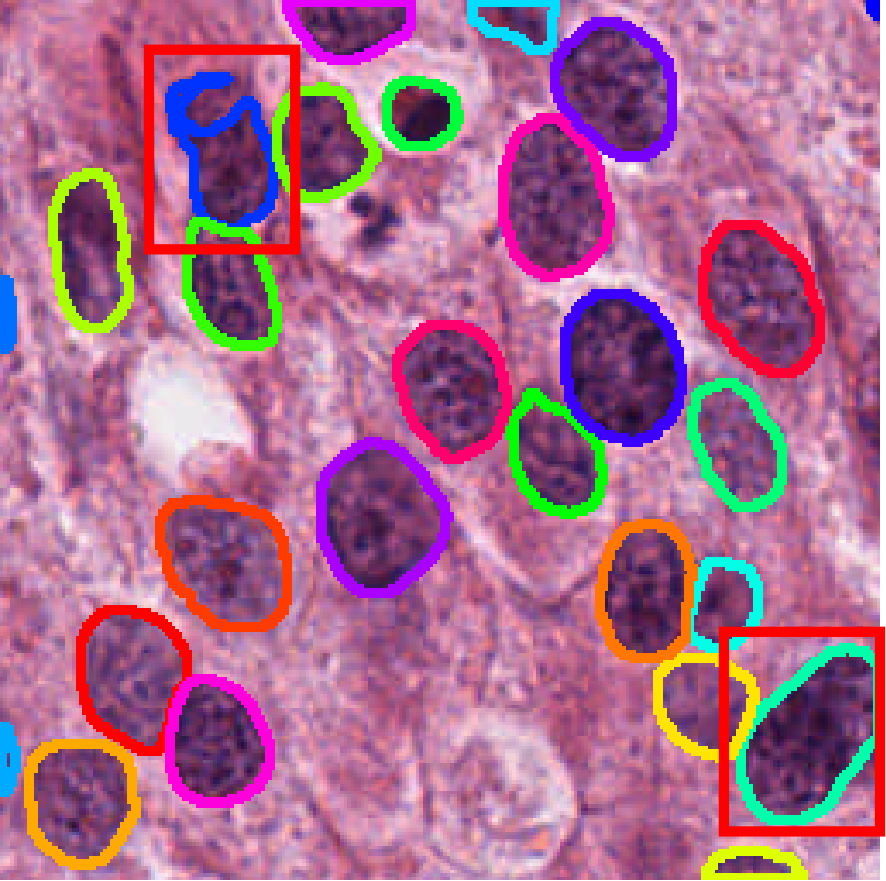} \\[1mm]

        \rotatebox{90}{Ours} &
        \includegraphics[width=\linewidth]{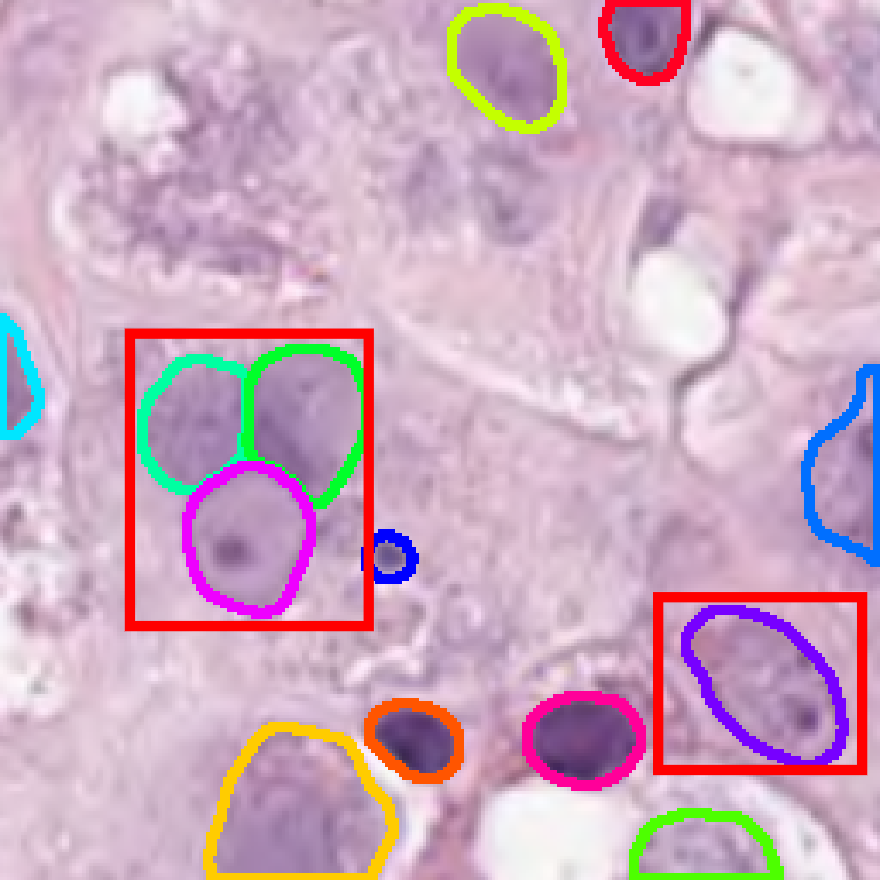} &
        \includegraphics[width=\linewidth]{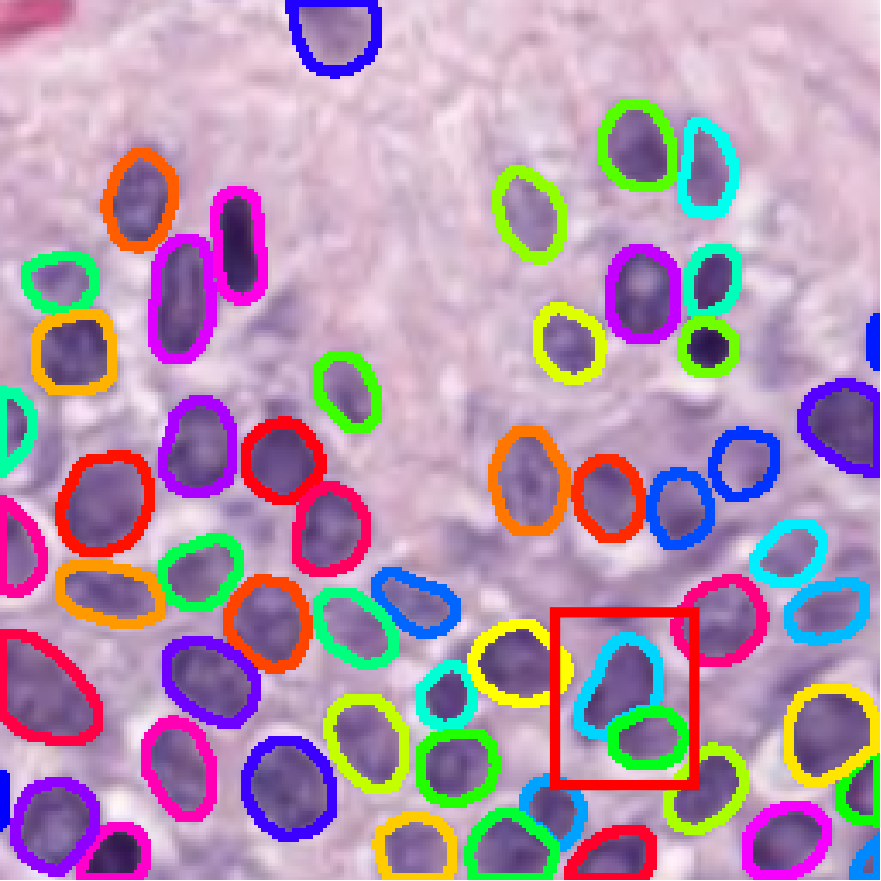} &
        \includegraphics[width=\linewidth]{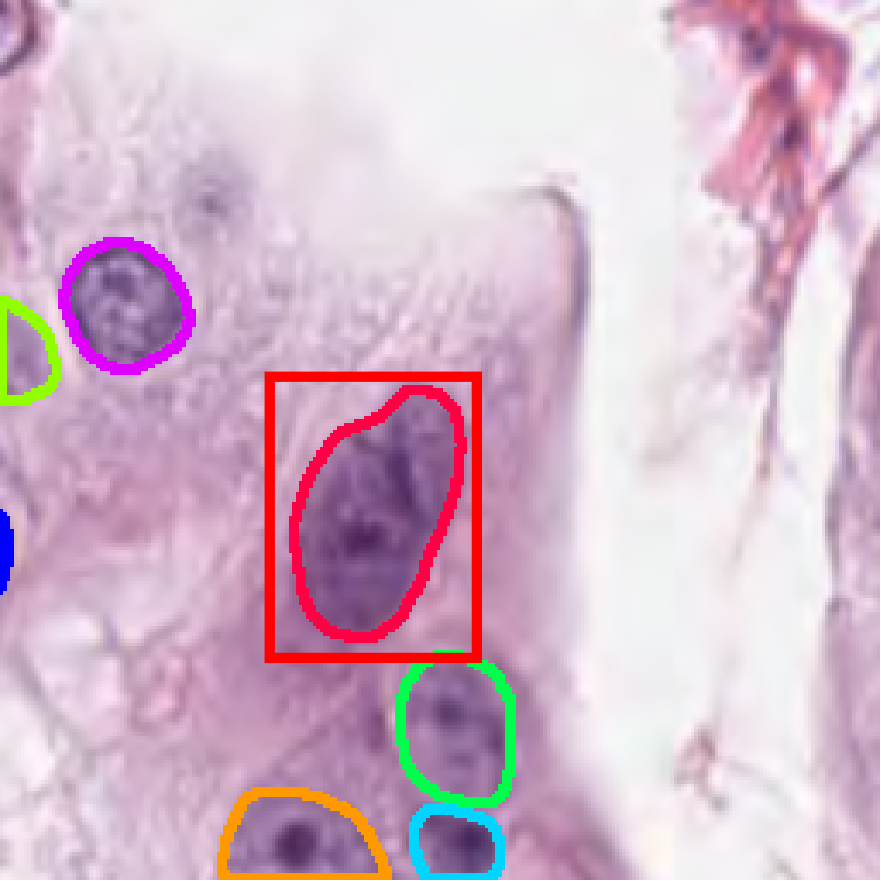} &
        \includegraphics[width=\linewidth]{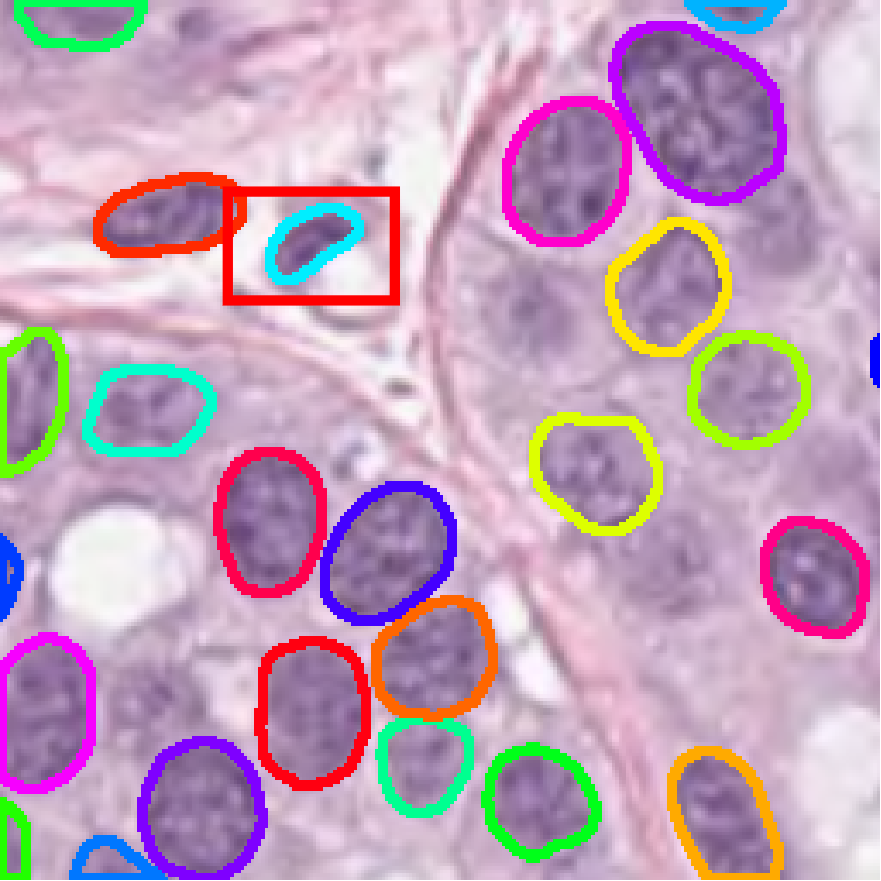} &
        \includegraphics[width=\linewidth]{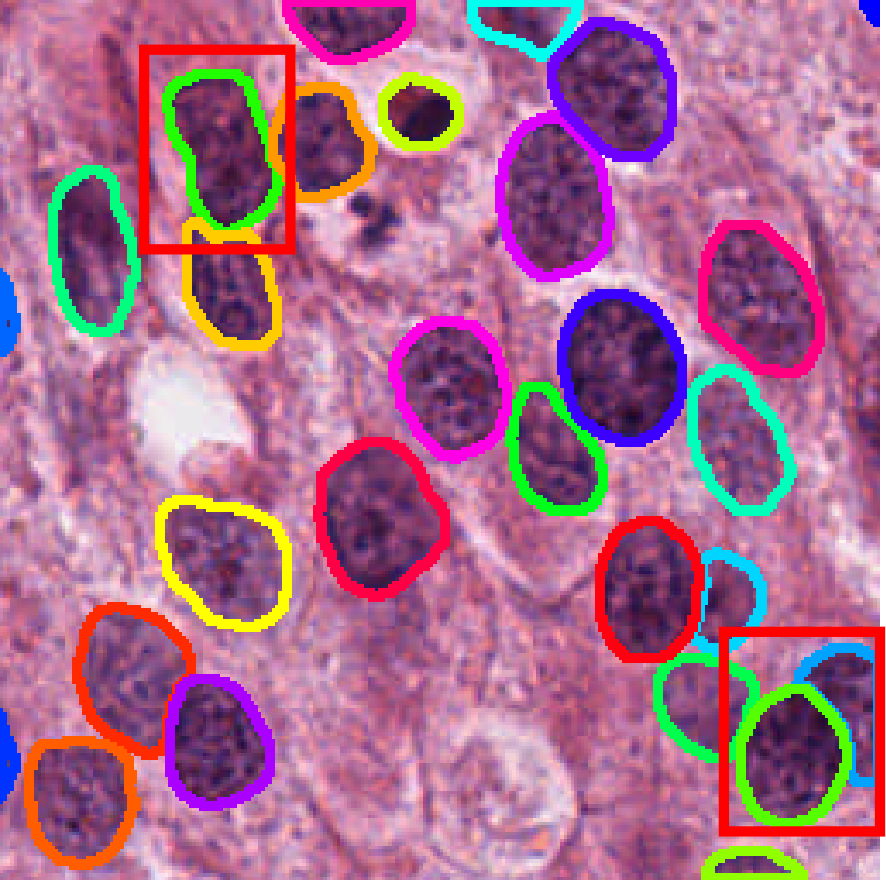} \\
    \end{tabular}
    
    \caption{Compared to the segmentation results of a full fine-tuned SAM (second row), our CFR-SAM achieves superior precision, particularly in capturing faint boundaries and distinguishing between densely clustered objects.}
    \label{fig:segmentation_comparison}
\end{figure}

%% file: refine.tex
\begin{figure}[t]
    \centering
    \footnotesize
    \begin{tabular}{@{}c@{\hspace{0.5mm}}c@{\hspace{0.5mm}}c@{\hspace{0.5mm}}c@{\hspace{0.5mm}}c@{}}
        \rotatebox{90}{\makebox[1.9cm][c]{GT}} &
        \includegraphics[width=0.11\textwidth,height=1.9cm]{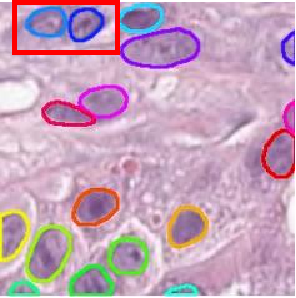} &
        \includegraphics[width=0.11\textwidth,height=1.9cm]{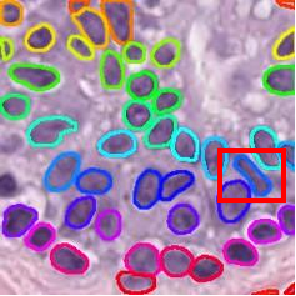} &
        \includegraphics[width=0.11\textwidth,height=1.9cm]{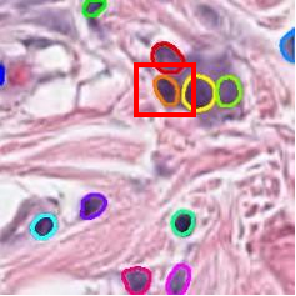} &
        \includegraphics[width=0.11\textwidth,height=1.9cm]{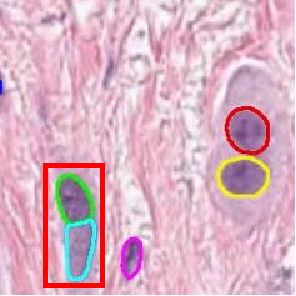} \\
        
        \rotatebox{90}{\makebox[1.9cm][c]{w/o BGMR}} &
        \includegraphics[width=0.11\textwidth,height=1.9cm]{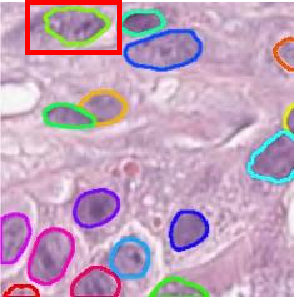} &
        \includegraphics[width=0.11\textwidth,height=1.9cm]{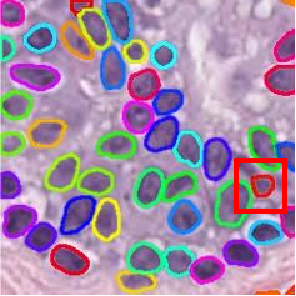} &
        \includegraphics[width=0.11\textwidth,height=1.9cm]{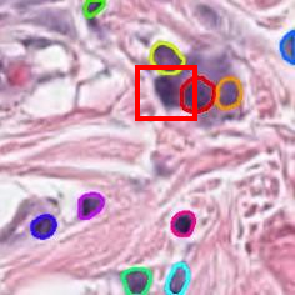} &
        \includegraphics[width=0.11\textwidth,height=1.9cm]{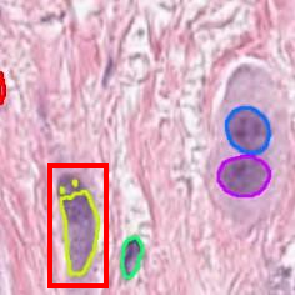} \\
        
        \rotatebox{90}{\makebox[1.9cm][c]{w/ BGMR}} &
        \includegraphics[width=0.11\textwidth,height=1.9cm]{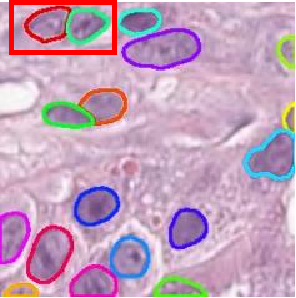} &
        \includegraphics[width=0.11\textwidth,height=1.9cm]{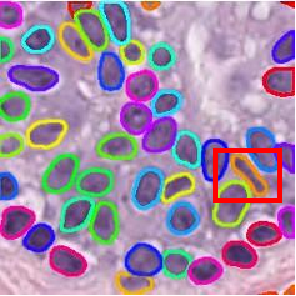} &
        \includegraphics[width=0.11\textwidth,height=1.9cm]{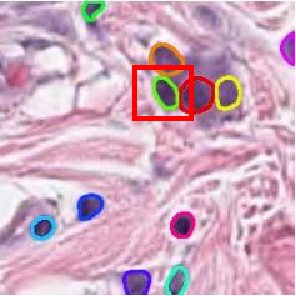} &
        \includegraphics[width=0.11\textwidth,height=1.9cm]{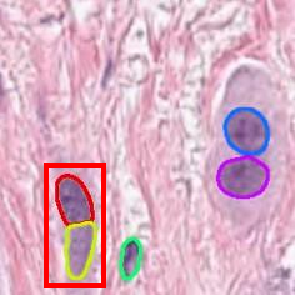} \\
    \end{tabular}
    
    \caption{Visual comparison of the BGMR ablation study for nuclei instance segmentation. Nuclear types are represented by distinct boundary colors.}
    \label{fig:ablation_refine}
\end{figure}

%% file: peft.tex
\begin{figure}[t]
    \centering
    \footnotesize
    \begin{tabular}{c@{}c@{}c@{}c@{}}
        \includegraphics[width=0.11\textwidth,height=1.9cm]{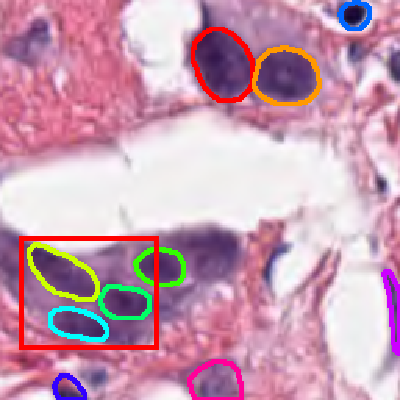} &
        \includegraphics[width=0.11\textwidth,height=1.9cm]{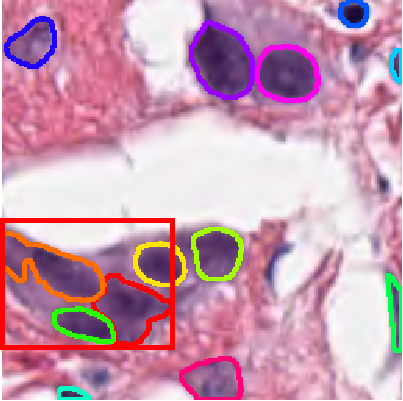} &
        \includegraphics[width=0.11\textwidth,height=1.9cm]{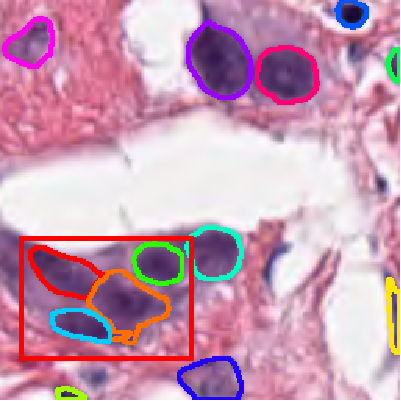} &
        \includegraphics[width=0.11\textwidth,height=1.9cm]{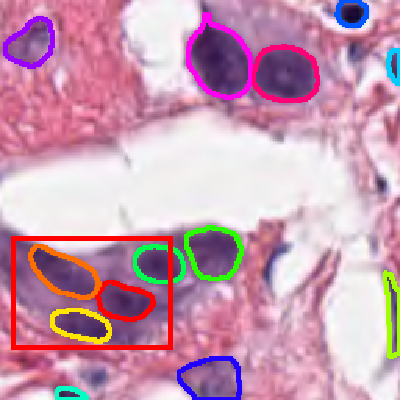} \\
        
        \includegraphics[width=0.11\textwidth,height=1.9cm]{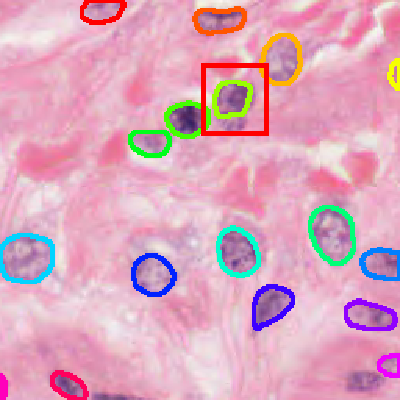} &
        \includegraphics[width=0.11\textwidth,height=1.9cm]{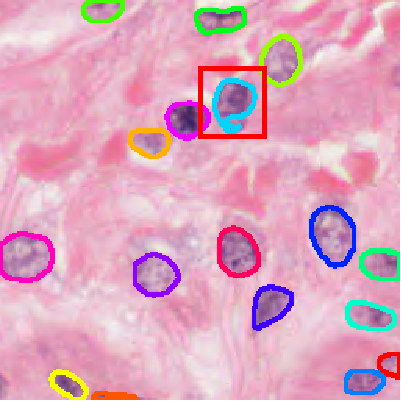} &
        \includegraphics[width=0.11\textwidth,height=1.9cm]{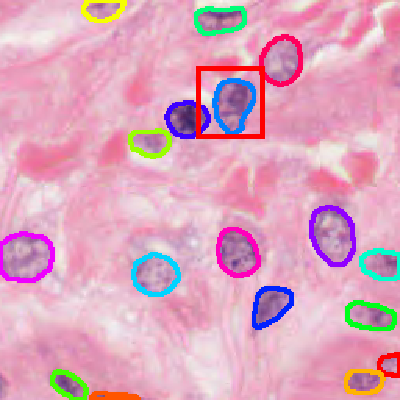} &
        \includegraphics[width=0.11\textwidth,height=1.9cm]{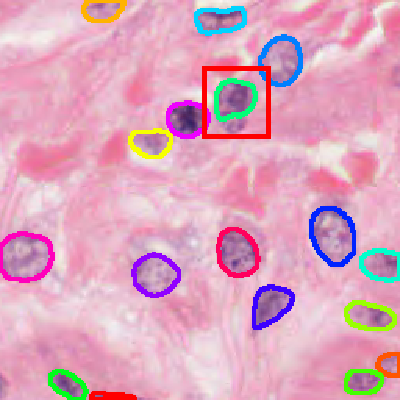} \\
        
        \includegraphics[width=0.11\textwidth,height=1.9cm]{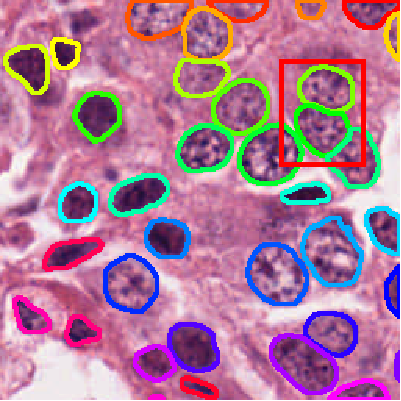} &
        \includegraphics[width=0.11\textwidth,height=1.9cm]{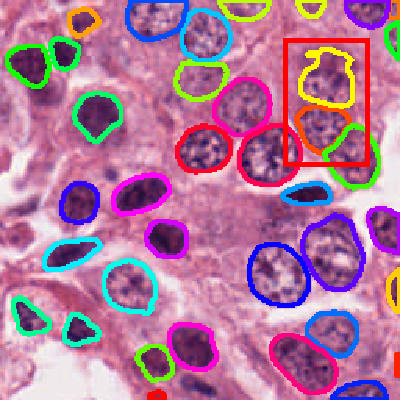} &
        \includegraphics[width=0.11\textwidth,height=1.9cm]{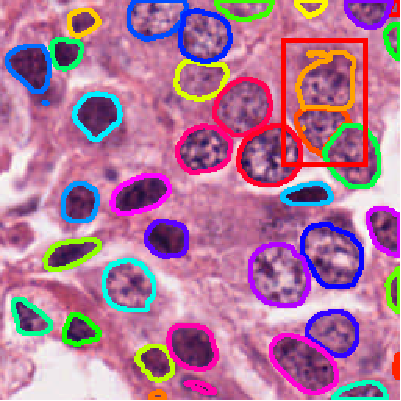} &
        \includegraphics[width=0.11\textwidth,height=1.9cm]{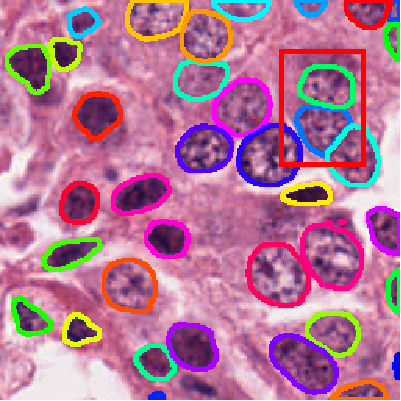} \\
        
        \multirow{2}{*}{GT} & \multirow{2}{*}{Adapter} & \multirow{2}{*}{LoRA} & \multirow{2}{*}{MALA} \\
        & & & \\
    \end{tabular}
    \caption{Visual comparison of different adapting strategies for nuclei instance segmentation.}
    \label{fig:ablation_peft}
\end{figure}

%% file: ablation_peft.tex
\definecolor{mygray}{gray}{.9}
\begin{table}[t]
\caption{Ablation study of different PEFT methods. It is important to note that these results are obtained using SAM adapted solely by the respective PEFT methods, without any other auxiliary modules. The symbol $*$ indicates that we increase the low-rank dimension of Adapter to approximately match the trainable parameters of MALA for a fair comparison.}
\renewcommand{\arraystretch}{1.1}
\centering
\resizebox{1\columnwidth}{!}{%
\begin{tabular}{lccc}
\toprule[2pt]
\multirow{2}{*}{Model}        & Trainable            & \multirow{2}{*}{mPQ} & \multirow{2}{*}{bPQ} \\
                               & Params (M)       &                      &                      \\ \midrule  
Adapter \citep{houlsby2019parameter}                       & 6.4                  & 0.4969               & 0.6799               \\
\review{Adapter$^{*}$} \citep{houlsby2019parameter}      &  \review{8.0}     &   \review{0.4955}  &   \review{0.6782}   \\
Lora\citep{hu2022lora}                           & 12.8                 & 0.4923               & 0.6741               \\
\rowcolor{gray!20} MALA                           & 7.8                  & 0.4983               & 0.6821               \\ \bottomrule[2pt]
\end{tabular}}

\label{tab:ablation_peft}
\end{table}

%% file: ablation_dimention.tex
\begin{table}[t]
\caption{Ablation study of different low-rank dimension.}
\renewcommand{\arraystretch}{1.1}
\centering
\begin{tabular}{lccc}
\toprule[2pt]
\multirow{2}{*}{Model}        & Trainable            & \multirow{2}{*}{mPQ} & \multirow{2}{*}{bPQ} \\
                               & Params (M)       &                      &                      \\ \midrule 
MALA$_{(16)}$                  & 5.6                  & 0.5033               & 0.6892               \\
MALA$_{(32)}$                  & 6.5                  & 0.5031               & 0.6889               \\
\rowcolor{gray!20} MALA$_{(64)}$                  & 8.3                  & 0.5046               & 0.6912     \\
MALA$_{(128)}$                 & 12.1                 & 0.5043               & 0.6911               \\ \bottomrule[2pt]
\end{tabular}

\label{tab:ablation_dimention}
\end{table}

%% file: mask_num.tex
\begin{table}[t]
\caption{Ablation study on the number of sampled nuclei per image during training.}
\renewcommand{\arraystretch}{1.1}
\centering
\setlength{\tabcolsep}{18pt}
\review{
\begin{tabular}{lcc}
\toprule[2pt]
Model           & mPQ & bPQ \\ \midrule 
Mask$_{(20)}$                        & 0.5043               & 0.6908               \\
\rowcolor{gray!20} Mask$_{(25)}$     & 0.5046               & 0.6912               \\
Mask$_{(30)}$                        & 0.5046               & 0.6910               \\
Mask$_{(35)}$                        & 0.5045               & 0.6906               \\ \bottomrule[2pt]
\end{tabular}
}
\label{tab:mask_num}
\end{table}

%% file: box.tex
\begin{table}[]
\renewcommand{\arraystretch}{1.1}
\centering
\caption{Comparison of point prompts and bounding box prompts for SAM adaptation.}
\review{
\begin{tabular}{lcc}
\toprule[2pt]
Model                   & mPQ & bPQ \\ \midrule 
Point (Ours-B)        & 0.5046      & 0.6912      \\                     
Box (Ours-B)          & 0.4804               & 0.6612               \\   
\bottomrule[2pt]
\end{tabular}
}
\label{tab:box}
\end{table}

%% file: efficient_compare.tex
\begin{table}[t]
\centering
\caption{Comparison of model parameters, computational cost and inference speed on the PanNuke dataset. The FPS of all methods is tested in a single NVIDIA RTX 3090 GPU.}
\renewcommand{\arraystretch}{1.1}
\resizebox{1\columnwidth}{!}{%
\begin{tabular}{lcccc}
\toprule[2pt]
\multirow{2}{*}{Model}      & Trainable & Total & \multirow{2}{*}{MACs (G)}   & \multirow{2}{*}{FPS}     \\ 
& Params (M) & Params (M) & & \\ \midrule
STARDIST \citep{schmidt2018cell}     & 122.8       & 122.8         & 263.6    & 17   \\
HoVer-Net \citep{graham2019hover}    & 37.6        & 37.6         & 150.0    & 7    \\
CPP-Net \citep{chen2023cpp}          & 122.8       & 122.8         & 264.4    & 14   \\
PointNu-Net \citep{yao2023pointnu}   & 158.1       & 158.1         & 335.1    & 11   \\
CellViT-B \citep{horst2024cellvit}   & 142.9       & 142.9         & 232.0    & 20   \\
Ours-B                               & 63.4        & 153.1         & 89.5     & 13     \\ \midrule

\review{CellViT-H \citep{horst2024cellvit}}   & \review{699.7}       & \review{699.7}         & \review{471.3}    & \review{13}   \\
\review{Ours-H}                               & \review{74.2}        & \review{650.5}         & \review{299.7}     & \review{7}     \\ 

\bottomrule[2pt]
\end{tabular}}

\label{tab:efficient}
\end{table}

%% file: reference.bib
@String(ECCV= {Eur. Conf. Comput. Vis.})

@String(ICLR = {Int. Conf. Learn. Represent.})

@String(ECCV  = {ECCV})

@String(ICLR  = {ICLR})

@article{graham2019hover,
  title={Hover-net: Simultaneous segmentation and classification of nuclei in multi-tissue histology images},
  author={Graham, Simon and Vu, Quoc Dang and Raza, Shan E Ahmed and Azam, Ayesha and Tsang, Yee Wah and Kwak, Jin Tae and Rajpoot, Nasir},
  journal={Medical image analysis},
  volume={58},
  year={2019},
  publisher={Elsevier}
}

@article{wu2025medical,
  title={Medical sam adapter: Adapting segment anything model for medical image segmentation},
  author={Wu, Junde and Wang, Ziyue and Hong, Mingxuan and Ji, Wei and Fu, Huazhu and Xu, Yanwu and Xu, Min and Jin, Yueming},
  journal={Medical image analysis},
  volume={102},
  pages={103547},
  year={2025},
  publisher={Elsevier}
}

@article{sirinukunwattana2018novel,
  title={Novel digital signatures of tissue phenotypes for predicting distant metastasis in colorectal cancer},
  author={Sirinukunwattana, Korsuk and Snead, David and Epstein, David and Aftab, Zia and Mujeeb, Imaad and Tsang, Yee Wah and Cree, Ian and Rajpoot, Nasir},
  journal={Scientific reports},
  volume={8},
  number={1},
  pages={13692},
  year={2018},
  publisher={Nature Publishing Group UK London}
}

@article{lu2018nuclear,
  title={Nuclear shape and orientation features from H\&E images predict survival in early-stage estrogen receptor-positive breast cancers},
  author={Lu, Cheng and Romo-Bucheli, David and Wang, Xiangxue and Janowczyk, Andrew and Ganesan, Shridar and Gilmore, Hannah and Rimm, David and Madabhushi, Anant},
  journal={Laboratory investigation},
  volume={98},
  number={11},
  pages={1438--1448},
  year={2018},
  publisher={Elsevier}
}

@article{elmore2015diagnostic,
  title={Diagnostic concordance among pathologists interpreting breast biopsy specimens},
  author={Elmore, Joann G and Longton, Gary M and Carney, Patricia A and Geller, Berta M and Onega, Tracy and Tosteson, Anna NA and Nelson, Heidi D and Pepe, Margaret S and Allison, Kimberly H and Schnitt, Stuart J and others},
  journal={Jama},
  volume={313},
  number={11},
  pages={1122--1132},
  year={2015},
  publisher={American Medical Association}
}

@article{dosovitskiy2020image,
  title={An image is worth 16x16 words: Transformers for image recognition at scale},
  author={Dosovitskiy, Alexey and Beyer, Lucas and Kolesnikov, Alexander and Weissenborn, Dirk and Zhai, Xiaohua and Unterthiner, Thomas and Dehghani, Mostafa and Minderer, Matthias and Heigold, Georg and Gelly, Sylvain and others},
  journal={arXiv preprint arXiv:2010.11929},
  year={2020}
}

@article{horst2024cellvit,
  title={Cellvit: Vision transformers for precise cell segmentation and classification},
  author={H{\"o}rst, Fabian and Rempe, Moritz and Heine, Lukas and Seibold, Constantin and Keyl, Julius and Baldini, Giulia and Ugurel, Selma and Siveke, Jens and Gr{\"u}nwald, Barbara and Egger, Jan and others},
  journal={Medical Image Analysis},
  volume={94},
  pages={103143},
  year={2024},
  publisher={Elsevier}
}

@article{ilyas2022tsfd,
  title={TSFD-Net: Tissue specific feature distillation network for nuclei segmentation and classification},
  author={Ilyas, Talha and Mannan, Zubaer Ibna and Khan, Abbas and Azam, Sami and Kim, Hyongsuk and De Boer, Friso},
  journal={Neural Networks},
  volume={151},
  pages={1--15},
  year={2022},
  publisher={Elsevier}
}

@inproceedings{kirillov2023segment,
  title={Segment anything},
  author={Kirillov, Alexander and Mintun, Eric and Ravi, Nikhila and Mao, Hanzi and Rolland, Chloe and Gustafson, Laura and Xiao, Tete and Whitehead, Spencer and Berg, Alexander C and Lo, Wan-Yen and others},
  booktitle={Proceedings of the IEEE/CVF international conference on computer vision},
  pages={4015--4026},
  year={2023}
}

@inproceedings{deng2025segment,
  title={Segment anything model (sam) for digital pathology: Assess zero-shot segmentation on whole slide imaging},
  author={Deng, Ruining and Cui, Can and Liu, Quan and Yao, Tianyuan and Remedios, Lucas W and Bao, Shunxing and Landman, Bennett A and Wheless, Lee E and Coburn, Lori A and Wilson, Keith T and others},
  booktitle={IS\&T International Symposium on Electronic Imaging},
  volume={37},
  pages={COIMG--132},
  year={2025}
}

@article{chen2024sam,
  title={Un-sam: Universal prompt-free segmentation for generalized nuclei images},
  author={Chen, Zhen and Xu, Qing and Liu, Xinyu and Yuan, Yixuan},
  journal={arXiv preprint arXiv:2402.16663},
  year={2024}
}

@inproceedings{xu2023sppnet,
  title={Sppnet: A single-point prompt network for nuclei image segmentation},
  author={Xu, Qing and Kuang, Wenwei and Zhang, Zeyu and Bao, Xueyao and Chen, Haoran and Duan, Wenting},
  booktitle={International Workshop on Machine Learning in Medical Imaging},
  pages={227--236},
  year={2023},
  organization={Springer}
}

@inproceedings{woo2018cbam,
  title={Cbam: Convolutional block attention module},
  author={Woo, Sanghyun and Park, Jongchan and Lee, Joon-Young and Kweon, In So},
  booktitle={Proceedings of the European conference on computer vision (ECCV)},
  pages={3--19},
  year={2018}
}

@article{ma2024segment,
  title={Segment anything in medical images},
  author={Ma, Jun and He, Yuting and Li, Feifei and Han, Lin and You, Chenyu and Wang, Bo},
  journal={Nature Communications},
  volume={15},
  number={1},
  pages={654},
  year={2024},
  publisher={Nature Publishing Group UK London}
}

@article{hu2022lora,
  title={Lora: Low-rank adaptation of large language models.},
  author={Hu, Edward J and Shen, Yelong and Wallis, Phillip and Allen-Zhu, Zeyuan and Li, Yuanzhi and Wang, Shean and Wang, Lu and Chen, Weizhu and others},
  journal={ICLR},
  volume={1},
  number={2},
  pages={3},
  year={2022}
}

@inproceedings{hatamizadeh2022unetr,
  title={Unetr: Transformers for 3d medical image segmentation},
  author={Hatamizadeh, Ali and Tang, Yucheng and Nath, Vishwesh and Yang, Dong and Myronenko, Andriy and Landman, Bennett and Roth, Holger R and Xu, Daguang},
  booktitle={Proceedings of the IEEE/CVF winter conference on applications of computer vision},
  pages={574--584},
  year={2022}
}

@article{wienert2012detection,
  title={Detection and segmentation of cell nuclei in virtual microscopy images: a minimum-model approach},
  author={Wienert, Stephan and Heim, Daniel and Saeger, Kai and Stenzinger, Albrecht and Beil, Michael and Hufnagl, Peter and Dietel, Manfred and Denkert, Carsten and Klauschen, Frederick},
  journal={Scientific reports},
  volume={2},
  number={1},
  pages={503},
  year={2012},
  publisher={Nature Publishing Group UK London}
}

@article{veta2013automatic,
  title={Automatic nuclei segmentation in H\&E stained breast cancer histopathology images},
  author={Veta, Mitko and Van Diest, Paul J and Kornegoor, Robert and Huisman, Andr{\'e} and Viergever, Max A and Pluim, Josien PW},
  journal={PloS one},
  volume={8},
  number={7},
  pages={e70221},
  year={2013},
  publisher={Public Library of Science San Francisco, USA}
}

@inproceedings{he2017mask,
  title={Mask r-cnn},
  author={He, Kaiming and Gkioxari, Georgia and Doll{\'a}r, Piotr and Girshick, Ross},
  booktitle={Proceedings of the IEEE international conference on computer vision},
  pages={2961--2969},
  year={2017}
}

@inproceedings{schmidt2018cell,
  title={Cell detection with star-convex polygons},
  author={Schmidt, Uwe and Weigert, Martin and Broaddus, Coleman and Myers, Gene},
  booktitle={International conference on medical image computing and computer-assisted intervention},
  pages={265--273},
  year={2018},
  organization={Springer}
}

@article{chen2023cpp,
  title={CPP-net: Context-aware polygon proposal network for nucleus segmentation},
  author={Chen, Shengcong and Ding, Changxing and Liu, Minfeng and Cheng, Jun and Tao, Dacheng},
  journal={IEEE Transactions on Image Processing},
  volume={32},
  pages={980--994},
  year={2023},
  publisher={IEEE}
}

@article{huang2024segment,
  title={Segment anything model for medical images?},
  author={Huang, Yuhao and Yang, Xin and Liu, Lian and Zhou, Han and Chang, Ao and Zhou, Xinrui and Chen, Rusi and Yu, Junxuan and Chen, Jiongquan and Chen, Chaoyu and others},
  journal={Medical Image Analysis},
  volume={92},
  pages={103061},
  year={2024},
  publisher={Elsevier}
}

@article{doan2022sonnet,
  title={SONNET: A self-guided ordinal regression neural network for segmentation and classification of nuclei in large-scale multi-tissue histology images},
  author={Doan, Tan NN and Song, Boram and Vuong, Trinh TL and Kim, Kyungeun and Kwak, Jin T},
  journal={IEEE Journal of Biomedical and Health Informatics},
  volume={26},
  number={7},
  pages={3218--3228},
  year={2022},
  publisher={IEEE}
}

@article{zhang2023customized,
  title={Customized segment anything model for medical image segmentation},
  author={Zhang, Kaidong and Liu, Dong},
  journal={arXiv preprint arXiv:2304.13785},
  year={2023}
}

@inproceedings{houlsby2019parameter,
  title={Parameter-efficient transfer learning for NLP},
  author={Houlsby, Neil and Giurgiu, Andrei and Jastrzebski, Stanislaw and Morrone, Bruna and De Laroussilhe, Quentin and Gesmundo, Andrea and Attariyan, Mona and Gelly, Sylvain},
  booktitle={International conference on machine learning},
  pages={2790--2799},
  year={2019},
  organization={PMLR}
}

@inproceedings{gamper2019pannuke,
  title={Pannuke: an open pan-cancer histology dataset for nuclei instance segmentation and classification},
  author={Gamper, Jevgenij and Alemi Koohbanani, Navid and Benet, Ksenija and Khuram, Ali and Rajpoot, Nasir},
  booktitle={European congress on digital pathology},
  pages={11--19},
  year={2019},
  organization={Springer}
}

@article{gamper2020pannuke,
  title={Pannuke dataset extension, insights and baselines},
  author={Gamper, Jevgenij and Koohbanani, Navid Alemi and Benes, Ksenija and Graham, Simon and Jahanifar, Mostafa and Khurram, Syed Ali and Azam, Ayesha and Hewitt, Katherine and Rajpoot, Nasir},
  journal={arXiv preprint arXiv:2003.10778},
  year={2020}
}

@article{vu2019methods,
  title={Methods for segmentation and classification of digital microscopy tissue images},
  author={Vu, Quoc Dang and Graham, Simon and Kurc, Tahsin and To, Minh Nguyen Nhat and Shaban, Muhammad and Qaiser, Talha and Koohbanani, Navid Alemi and Khurram, Syed Ali and Kalpathy-Cramer, Jayashree and Zhao, Tianhao and others},
  journal={Frontiers in bioengineering and biotechnology},
  volume={7},
  pages={53},
  year={2019},
  publisher={Frontiers Media SA}
}

@article{kumar2019multi,
  title={A multi-organ nucleus segmentation challenge},
  author={Kumar, Neeraj and Verma, Ruchika and Anand, Deepak and Zhou, Yanning and Onder, Omer Fahri and Tsougenis, Efstratios and Chen, Hao and Heng, Pheng-Ann and Li, Jiahui and Hu, Zhiqiang and others},
  journal={IEEE transactions on medical imaging},
  volume={39},
  number={5},
  pages={1380--1391},
  year={2019},
  publisher={IEEE}
}

@inproceedings{nam2024instasam,
  title={Instasam: Instance-aware segment any nuclei model with point annotations},
  author={Nam, Siwoo and Namgung, Hyun and Jeong, Jaehoon and Luna, Miguel and Kim, Soopil and Chikontwe, Philip and Park, Sang Hyun},
  booktitle={International Conference on Medical Image Computing and Computer-Assisted Intervention},
  pages={232--242},
  year={2024},
  organization={Springer}
}

@article{naylor2018segmentation,
  title={Segmentation of nuclei in histopathology images by deep regression of the distance map},
  author={Naylor, Peter and La{\'e}, Marick and Reyal, Fabien and Walter, Thomas},
  journal={IEEE transactions on medical imaging},
  volume={38},
  number={2},
  pages={448--459},
  year={2018},
  publisher={IEEE}
}

@article{raza2019micro,
  title={Micro-Net: A unified model for segmentation of various objects in microscopy images},
  author={Raza, Shan E Ahmed and Cheung, Linda and Shaban, Muhammad and Graham, Simon and Epstein, David and Pelengaris, Stella and Khan, Michael and Rajpoot, Nasir M},
  journal={Medical image analysis},
  volume={52},
  pages={160--173},
  year={2019},
  publisher={Elsevier}
}

@inproceedings{ronneberger2015u,
  title={U-net: Convolutional networks for biomedical image segmentation},
  author={Ronneberger, Olaf and Fischer, Philipp and Brox, Thomas},
  booktitle={International Conference on Medical image computing and computer-assisted intervention},
  pages={234--241},
  year={2015},
  organization={Springer}
}

@article{chen2017dcan,
  title={DCAN: Deep contour-aware networks for object instance segmentation from histology images},
  author={Chen, Hao and Qi, Xiaojuan and Yu, Lequan and Dou, Qi and Qin, Jing and Heng, Pheng-Ann},
  journal={Medical image analysis},
  volume={36},
  pages={135--146},
  year={2017},
  publisher={Elsevier}
}

@article{deshmukh2022feednet,
  title={FEEDNet: A feature enhanced encoder-decoder LSTM network for nuclei instance segmentation for histopathological diagnosis},
  author={Deshmukh, Gayatri and Susladkar, Onkar and Makwana, Dhruv and Mittal, Sparsh and others},
  journal={Physics in Medicine \& Biology},
  volume={67},
  number={19},
  pages={195011},
  year={2022},
  publisher={IOP Publishing}
}

@article{chen2023enhancing,
  title={Enhancing nucleus segmentation with haru-net: a hybrid attention based residual u-blocks network},
  author={Chen, Junzhou and Huang, Qian and Chen, Yulin and Qian, Linyi and Yu, Chengyuan},
  journal={arXiv preprint arXiv:2308.03382},
  year={2023}
}

@article{yao2023pointnu,
  title={Pointnu-net: Keypoint-assisted convolutional neural network for simultaneous multi-tissue histology nuclei segmentation and classification},
  author={Yao, Kai and Huang, Kaizhu and Sun, Jie and Hussain, Amir},
  journal={IEEE Transactions on Emerging Topics in Computational Intelligence},
  volume={8},
  number={1},
  pages={802--813},
  year={2023},
  publisher={IEEE}
}

@inproceedings{zhou2019cia,
  title={Cia-net: Robust nuclei instance segmentation with contour-aware information aggregation},
  author={Zhou, Yanning and Onder, Omer Fahri and Dou, Qi and Tsougenis, Efstratios and Chen, Hao and Heng, Pheng-Ann},
  booktitle={International conference on information processing in medical imaging},
  pages={682--693},
  year={2019},
  organization={Springer}
}

@inproceedings{he2021cdnet,
  title={Cdnet: Centripetal direction network for nuclear instance segmentation},
  author={He, Hongliang and Huang, Zhongyi and Ding, Yao and Song, Guoli and Wang, Lin and Ren, Qian and Wei, Pengxu and Gao, Zhiqiang and Chen, Jie},
  booktitle={Proceedings of the IEEE/CVF International Conference on Computer Vision},
  pages={4026--4035},
  year={2021}
}

@inproceedings{he2023toposeg,
  title={Toposeg: Topology-aware nuclear instance segmentation},
  author={He, Hongliang and Wang, Jun and Wei, Pengxu and Xu, Fan and Ji, Xiangyang and Liu, Chang and Chen, Jie},
  booktitle={Proceedings of the IEEE/CVF International Conference on Computer Vision},
  pages={21307--21316},
  year={2023}
}

@article{kuhn1955hungarian,
  title={The Hungarian method for the assignment problem},
  author={Kuhn, Harold W},
  journal={Naval research logistics quarterly},
  volume={2},
  number={1-2},
  pages={83--97},
  year={1955},
  publisher={Wiley Online Library}
}

@inproceedings{song2021rethinking,
  title={Rethinking counting and localization in crowds: A purely point-based framework},
  author={Song, Qingyu and Wang, Changan and Jiang, Zhengkai and Wang, Yabiao and Tai, Ying and Wang, Chengjie and Li, Jilin and Huang, Feiyue and Wu, Yang},
  booktitle={Proceedings of the IEEE/CVF international conference on computer vision},
  pages={3365--3374},
  year={2021}
}

@inproceedings{lin2017focal,
  title={Focal loss for dense object detection},
  author={Lin, Tsung-Yi and Goyal, Priya and Girshick, Ross and He, Kaiming and Doll{\'a}r, Piotr},
  booktitle={Proceedings of the IEEE international conference on computer vision},
  pages={2980--2988},
  year={2017}
}

@article{li2025nuhtc,
  title={NuHTC: A hybrid task cascade for nuclei instance segmentation and classification},
  author={Li, Bao and Liu, Zhenyu and Zhang, Song and Liu, Xiangyu and Sun, Caixia and Liu, Jiangang and Qiu, Bensheng and Tian, Jie},
  journal={Medical Image Analysis},
  pages={103595},
  year={2025},
  publisher={Elsevier}
}

@inproceedings{konwer2025enhancing,
  title={Enhancing SAM with efficient prompting and preference optimization for semi-supervised medical image segmentation},
  author={Konwer, Aishik and Yang, Zhijian and Bas, Erhan and Xiao, Cao and Prasanna, Prateek and Bhatia, Parminder and Kass-Hout, Taha},
  booktitle={Proceedings of the IEEE/CVF Conference on Computer Vision and Pattern Recognition},
  pages={20990--21000},
  year={2025}
}

@inproceedings{spiegler2025textsam,
  title={Textsam-eus: Text prompt learning for sam to accurately segment pancreatic tumor in endoscopic ultrasound},
  author={Spiegler, Pascal and Koleilat, Taha and Harirpoush, Arash and Miller, Corey S and Rivaz, Hassan and Kersten-Oertel, Marta and Xiao, Yiming},
  booktitle={Proceedings of the IEEE/CVF International Conference on Computer Vision},
  pages={948--957},
  year={2025}
}

@inproceedings{xiao2024cat,
  title={Cat-sam: Conditional tuning for few-shot adaptation of segment anything model},
  author={Xiao, Aoran and Xuan, Weihao and Qi, Heli and Xing, Yun and Ren, Ruijie and Zhang, Xiaoqin and Shao, Ling and Lu, Shijian},
  booktitle={European Conference on Computer Vision},
  pages={189--206},
  year={2024},
  organization={Springer}
}

@inproceedings{paranjape2024s,
  title={S-sam: Svd-based fine-tuning of segment anything model for medical image segmentation},
  author={Paranjape, Jay N and Sikder, Shameema and Vedula, S Swaroop and Patel, Vishal M},
  booktitle={International Conference on Medical Image Computing and Computer-Assisted Intervention},
  pages={720--730},
  year={2024},
  organization={Springer}
}

@article{koleilat2024medclip,
  title={Medclip-samv2: Towards universal text-driven medical image segmentation. arXiv 2024},
  author={Koleilat, T and Asgariandehkordi, H and Rivaz, H and Xiao, Y},
  journal={arXiv preprint arXiv:2409.19483},
  year={2024}
}

@inproceedings{paranjape2024adaptivesam,
  title={Adaptivesam: Towards efficient tuning of sam for surgical scene segmentation},
  author={Paranjape, Jay N and Nair, Nithin Gopalakrishnan and Sikder, Shameema and Vedula, S Swaroop and Patel, Vishal M},
  booktitle={Annual Conference on Medical Image Understanding and Analysis},
  pages={187--201},
  year={2024},
  organization={Springer}
}

@article{tommasino2026nulite,
  title={Nulite-lightweight and fast model for nuclei instance segmentation and classification},
  author={Tommasino, Cristian and Russo, Cristiano and Rinaldi, Antonio M},
  journal={Biomedical Signal Processing and Control},
  volume={114},
  pages={109333},
  year={2026},
  publisher={Elsevier}
}

@inproceedings{tommasino2024hover,
  title={" HoVer-UNet": Accelerating Hovernet with Unet-Based Multi-Class Nuclei Segmentation Via Knowledge Distillation},
  author={Tommasino, Cristian and Russo, Cristiano and Rinaldi, Antonio Maria and Ciompi, Francesco},
  booktitle={2024 IEEE International symposium on biomedical imaging (ISBI)},
  pages={1--4},
  year={2024},
  organization={IEEE}
}

@article{kumar2017dataset,
  title={A dataset and a technique for generalized nuclear segmentation for computational pathology},
  author={Kumar, Neeraj and Verma, Ruchika and Sharma, Sanuj and Bhargava, Surabhi and Vahadane, Abhishek and Sethi, Amit},
  journal={IEEE transactions on medical imaging},
  volume={36},
  number={7},
  pages={1550--1560},
  year={2017},
  publisher={IEEE}
}

@article{corbetta2025multi,
  title={Multi-marker similarity enables reduced-reference and interpretable image quality assessment in optical microscopy},
  author={Corbetta, Elena and Bocklitz, Thomas},
  journal={Research},
  volume={8},
  pages={0783},
  year={2025},
  publisher={AAAS}
}

@article{ding2025artificial,
  title={Artificial Intelligence for Organelle Segmentation in Live-Cell Imaging},
  author={Ding, Yang and Tan, Zhijun and Li, Jintao and Zhang, Weisen and Fang, Bin and Bai, Hua and Xin, Weini and Voelcker, Nicolas H and Peng, Bo and Li, Lin},
  journal={Research},
  volume={8},
  pages={1035},
  year={2025},
  publisher={AAAS}
}

@article{wu2025segnet,
  title={E-SegNet: E-Shaped Structure Networks for Accurate 2D and 3D Medical Image Segmentation},
  author={Wu, Wei and Yang, Xin and Yao, Chenggui and Liu, Ou and Zhao, Qi and Shuai, Jianwei},
  journal={Research},
  volume={8},
  pages={0869},
  year={2025},
  publisher={AAAS}
}

@article{wang2024masked,
  title={Masked Generative Light Field Prompting for Pixel-Level Structure Segmentations},
  author={Wang, Mianzhao and Shi, Fan and Cheng, Xu and Chen, Shengyong},
  journal={Research},
  volume={7},
  pages={0328},
  year={2024},
  publisher={AAAS}
}

@article{jiao2026foundation,
  title={Foundation Models Meet Medical Image Interpretation},
  author={Jiao, Licheng and Hao, Jiayao and Li, Ruiyang and Li, Lingling and Liu, Xu and Liu, Fang and Ma, Wenping and Chen, Puhua and Huang, Zhongjian and Yang, Jingyi and others},
  journal={Research},
  volume={9},
  pages={1024},
  year={2026},
  publisher={AAAS}
}
